\pdfoutput=1
\documentclass{article}

    \PassOptionsToPackage{numbers, compress}{natbib}

\usepackage[preprint]{neurips_data_2023}

\usepackage[utf8]{inputenc} 
\usepackage[T1]{fontenc}    
\usepackage{hyperref}       
\usepackage{url}            
\usepackage{booktabs}       
\usepackage{amsfonts}       
\usepackage{nicefrac}       
\usepackage{microtype}      
\usepackage{xcolor}         

\usepackage{graphicx}
\usepackage{subfigure}
\usepackage{booktabs} 
\usepackage{makecell}
\usepackage{multirow}
\usepackage{listings}
\usepackage{pythonhighlight}
\usepackage{wrapfig}

\usepackage{amsmath,amsfonts,bm}

\def\Acal{{\mathcal{A}}}

\def\Mcal{{\mathcal{M}}}

\def\Pcal{{\mathcal{P}}}

\def\Scal{{\mathcal{S}}}

\def\eqref#1{equation~\ref{#1}}

\def\1{\bm{1}}

\DeclareMathAlphabet{\mathsfit}{\encodingdefault}{\sfdefault}{m}{sl}
\SetMathAlphabet{\mathsfit}{bold}{\encodingdefault}{\sfdefault}{bx}{n}

\usepackage[toc,page]{appendix}
\usepackage{minitoc}

\title{Datasets and Benchmarks for Offline Safe Reinforcement Learning}

\author{%
  Zuxin Liu$^{1}$\thanks{equal contribution, corresponding to: \texttt{zuxinl@andrew.cmu.edu}},
  Zijian Guo$^{1*}$, Haohong Lin$^1$, Yihang Yao$^1$, Jiacheng Zhu$^1$, Zhepeng Cen$^1$,  \\
  \textbf{Hanjiang Hu$^1$, Wenhao Yu$^2$, Tingnan Zhang$^2$,
  Jie Tan$^2$, 
  Ding Zhao$^1$} \\
  $^1$ Carnegie Mellon University, $^2$ Google Deepmind
}

\begin{document}

\maketitle

\begin{abstract}
  This paper presents a comprehensive benchmarking suite tailored to offline safe reinforcement learning (RL) challenges, aiming to foster progress in the development and evaluation of safe learning algorithms in both the training and deployment phases. 
  Our benchmark suite contains three packages: 1) expertly crafted safe policies, 2) D4RL-styled datasets along with environment wrappers, and 3) high-quality offline safe RL baseline implementations.
  We feature a methodical data collection pipeline powered by advanced safe RL algorithms, which facilitates the generation of diverse datasets across 38 popular safe RL tasks, from robot control to autonomous driving.
  We further introduce an array of data post-processing filters, capable of modifying each dataset's diversity, thereby simulating various data collection conditions. 
  Additionally, we provide elegant and extensible implementations of prevalent offline safe RL algorithms to accelerate research in this area. Through extensive experiments with over 50000 CPU and 800 GPU hours of computations, we evaluate and compare the performance of these baseline algorithms on the collected datasets, offering insights into their strengths, limitations, and potential areas of improvement. Our benchmarking framework serves as a valuable resource for researchers and practitioners, facilitating the development of more robust and reliable offline safe RL solutions in safety-critical applications. The benchmark website is available at \url{www.offline-saferl.org}.
\end{abstract}



\section{Introduction}
\label{sec:intro}
Reinforcement learning (RL) has shown remarkable success in a broad array of domains, from game playing to robot control~\cite{mnih2015human, ibarz2021train}. Nevertheless, a paramount challenge persists: guaranteeing safety throughout both the training and deployment phases~\cite{gu2022review}. This is of particular concern in applications where unsafe behaviors could lead to catastrophic outcomes~\cite{xu2022trustworthy}. 
As a result, offline safe learning  emerges as an essential area of research, aiming to learn constrained policies from pre-collected datasets that satisfy safety requirements throughout the learning process~\cite{levine2020offline}.

\begin{figure}[t]
    \centering    \includegraphics[width=0.95\linewidth]{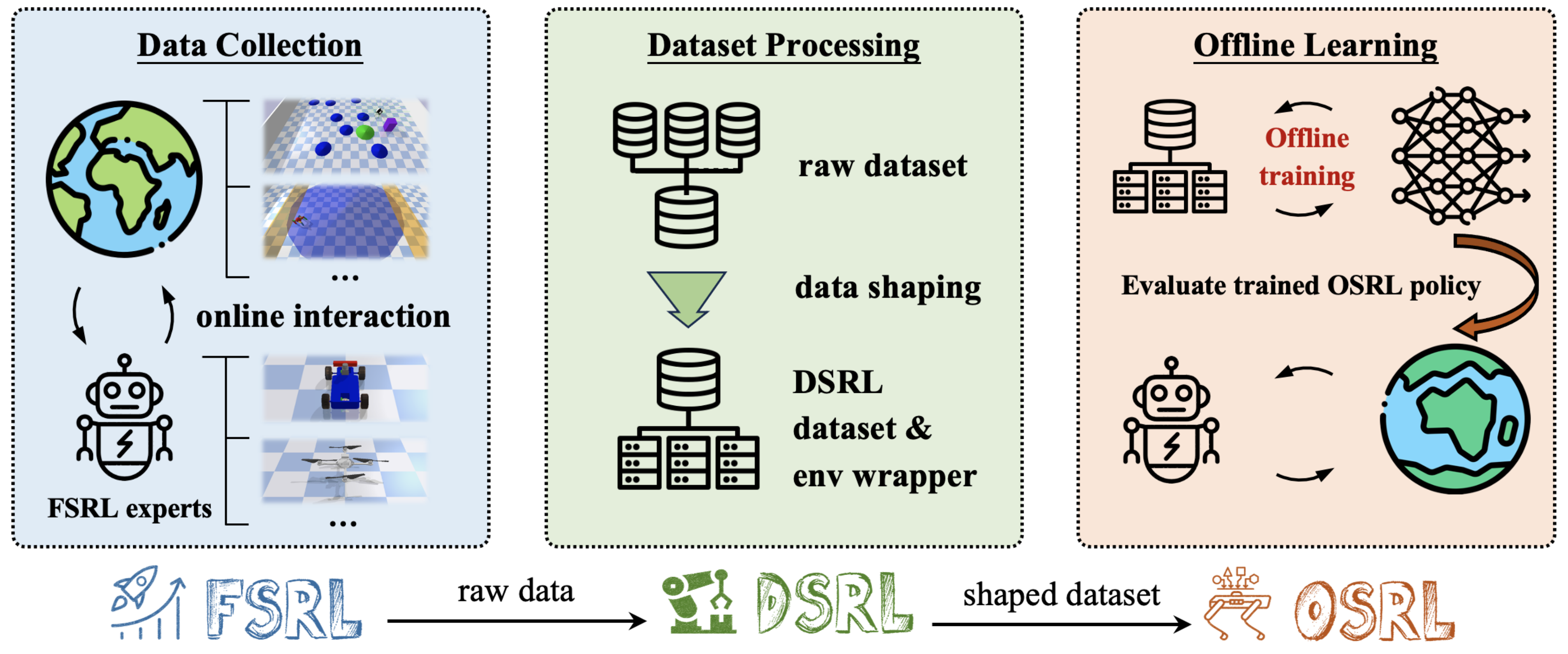}
    \caption{Overview of the benchmarking platform and the three packages: \texttt{FSRL}, \texttt{DSRL}, and \texttt{OSRL}.}
    \label{fig:overview}
    \vspace{-5mm}
\end{figure}

Despite the rising significance of offline learning, public benchmarks or datasets specifically designed to address the safety aspect are notably scarce. Conventional datasets, like D4RL~\cite{fu2020d4rl}, are excellent proving grounds for traditional offline learning algorithms, but their primary objective is reward maximization without any explicit safety constraints~\cite{seno2022d3rlpy}. This notable gap impedes the progress of deploying RL safely to real-world applications. There is a clear need for a specialized benchmark and dataset to train, evaluate, and compare safe learning algorithms under constraints.

To address this gap, we introduce a comprehensive benchmarking platform and datasets for offline safe learning, comprised of three packages: \href{https://github.com/liuzuxin/FSRL}{\texttt{FSRL}} (Fast Safe RL), \href{https://github.com/liuzuxin/DSRL}{\texttt{DSRL}} (Datasets for Safe RL), and \href{https://github.com/liuzuxin/OSRL}{\texttt{OSRL}} (Offline Safe RL) as shown in Figure \ref{fig:overview}. \texttt{FSRL} incorporates an efficient data collection pipeline, leveraging fast implementations of advanced safe RL algorithms to generate datasets suitable for safety-embedded tasks. 
We provide 38 high-quality datasets across different difficulty levels in three widely-used safe RL environments: Mujoco-based SafetyGymnasium~\cite{ray2019benchmarking, ji2023omnisafe}, PyBullet-based BulletSafetyGym~\cite{gronauer2022bullet}, and Panda3D-based self-driving simulator MetaDrive~\cite{li2022metadrive}.

\texttt{DSRL} hosts these datasets, offering a consistent API with D4RL for easy usage and evaluation of offline learning methods~\cite{fu2020d4rl}. We also furnish an array of deterministic data post-processing filters that can alter data density, noise level, as well as the distributions of rewards and costs, simulating diverse data collection conditions. This produces hundreds of distinct datasets of varying difficulty levels. Importantly, our framework doesn't only supply pre-collected datasets but also establishes a systematic approach to data collection and processing, enabling easy extension to other domains for future datasets and fostering a continually evolving benchmarking ecosystem.

Furthermore, we offer the \texttt{OSRL} codebase, implementing a broad spectrum of existing offline safe learning algorithms~\cite{xu2022constraints, lee2022coptidice, liu2023constrained}. This serves as a solid foundation for the safe RL community to build upon and benchmark against. To provide insights into their strengths and limitations, we conduct an extensive empirical analysis of these baseline algorithms using our benchmark datasets.

In summary, our contributions are as follows:

\noindent \textbf{1. We introduce a comprehensive benchmarking platform tailored for offline safe learning}, providing a standard testing ground for the evaluation and comparison of safe learning algorithms.

\noindent \textbf{2. We develop a data collection pipeline with advanced safe RL methods and filters}, generating diverse datasets across three environments with varying difficulty levels.


\noindent \textbf{3. We implement the D4RL-style data wrapper and many offline safe learning algorithms}, serving as a good starting point for researchers and practitioners in this area.

\noindent \textbf{4. We conduct a thorough empirical analysis}, utilizing over 50,000 CPU hours and 800 GPU hours of computation, providing us insights into the strengths and limitations of offline safe RL algorithms.

By making our \href{https://github.com/liuzuxin/DSRL}{datasets} and codebase publicly available, we aim to foster collaboration, accelerate innovation, and contribute to the broader adoption of safe RL solutions in safety-critical applications.

\vspace{-1mm}
\section{Related Work}
\vspace{-1mm}
\label{sec:related}
\textbf{Safe RL and Benchmarks.} Ensuring safety during RL training and deployment is a challenging problem~\cite{ gu2022review, xu2022trustworthy}. Numerous techniques have been explored to incorporate safety constraints into RL, such as constrained optimization~\cite{sootla2022saute, yang2021wcsac, liu2022constrained}, Lagrangian-based methods~\cite{chow2017risk, chen2021primal}, and correction-based approaches~\cite{zhao2021model, luo2021learning}. Despite these efforts, guaranteeing zero constraint violations during training is a formidable task~\cite{dalal2018safe, brunke2021safe}.
While there are benchmarks available for  safe RL \cite{ray2019benchmarking, gronauer2022bullet, ji2023omnisafe}, the lack of a comprehensive suite that targets offline training remains a gap in the field.

\textbf{Offline RL and Benchmarks.} Offline RL techniques aim to learn effective policies from pre-collected data without further environment interactions~\cite{ernst2005tree, levine2020offline}. It promises to enhance the scalability and efficiency of RL, particularly in applications where real-time interaction is expensive, risky, or impractical~\cite{fu2019diagnosing, brandfonbrener2021offline}. Offline RL is characterized by unique challenges, primarily arising from distributional shift, which can lead to extrapolation errors when learning policies beyond the support of the data distribution~\cite{fujimoto2019off, kumar2020conservative}. To combat these challenges, several strategies have been proposed, such as incorporating regularization or constraints~\cite{wu2019behavior, peng2019advantage, kostrikov2021offline}, or leveraging techniques like importance sampling to reduce estimation variance~\cite{nachum2019dualdice}.
While there are prevalent testing grounds for offline RL algorithms \cite{fu2020d4rl, gulcehre2020rl, tarasov2022corl}; however, they lack explicit safety constraints in their datasets.

\textbf{Offline Safe RL.} The intersection of offline RL and safe RL has recently been a focus of attention, where techniques from both fields are leveraged~\cite{le2019batch}. For instance, stationary distribution correction-based methods have been used to formulate the constrained optimization problem~\cite{lee2022coptidice, polosky2022constrained}, while Lagrangian-based approaches have been integrated with offline RL methods to provide safe learning~\cite{xu2022constraints}.
Sequential decision-making algorithms such as Decision Transformers have also been explored in this area \cite{liu2023constrained, zhang2023saformer}.
Despite these developments, there are no public safe RL datasets and algorithm libraries to compare these methods, revealing a clear need for a benchmarking framework to promote further research in this vital area.

\vspace{-1mm}
\section{Preliminaries}
\label{sec:preliminary}
\subsection{Constrained Markov Decision Process and Safe RL}

Safe RL is usually formulated under the Constrained Markov Decision Process (CMDP) framework~\cite{altman1998constrained}. A finite horizon CMDP, denoted as $\Mcal$, consists of a tuple $(\Scal, \Acal, \Pcal, r, c, \mu_0)$, where $\Scal$ represents the state space, $\Acal$ the action space, $\Pcal: \Scal \times \Acal \times \Scal \rightarrow [0,1]$ the transition function, $r: \Scal \times \Acal \times \Scal \rightarrow \mathbb{R}$ the reward function, and $\mu_0: \Scal \rightarrow [0, 1]$ the initial state distribution. In addition to these elements in a typical MDP, CMDP incorporates an extra cost function $c: \Scal \times \Acal \times \Scal \rightarrow [0, C_{max}] $ to account for constraint violations, with $C_{max}$ being the maximum cost.

A safe RL problem is specified by a CMDP and a constraint threshold $\kappa$. A policy $\pi: \Scal \times \Acal \rightarrow [0,1]$ maps the state-action space to probabilities, and a trajectory $\tau = \{s_1, a_1, r_1, c_1 ..., s_T, a_T, r_T, c_T \}$ contains state and action, reward, and cost information throughout the maximum episode length $T$. The cumulative reward and cost for a trajectory $\tau$ are represented as $R(\tau) = \sum_{t=1}^{T} r_t$ and $C(\tau)= \sum_{t=1}^{T} c_t$, respectively. Safe RL aims to find a policy that maximizes the reward while keeping the constraint violation cost below the threshold $\kappa$:
\begin{equation}
   \max_{\pi} \mathbb{E}_{\tau \sim \pi}  \big[R(\tau)], \quad s.t. \quad \mathbb{E}_{\tau \sim \pi} \big[C(\tau)] \leq \kappa. 
   \label{eq:safe-rl}
\end{equation}
The majority of literature focuses on the online setting \cite{achiam2017constrained, zhang2020first, liu2022robustness}, where the agent is allowed to interact with the environment to gather fresh trajectory data. Conversely, in the offline setting, the agent must rely on pre-collected trajectories from unknown policies, which poses challenges for solving this constrained optimization problem.

\subsection{Characterizing Dataset with Constraints}
\label{sec:offline-safe-rl}

\begin{figure}[t]
    \centering    \includegraphics[width=0.99\linewidth]{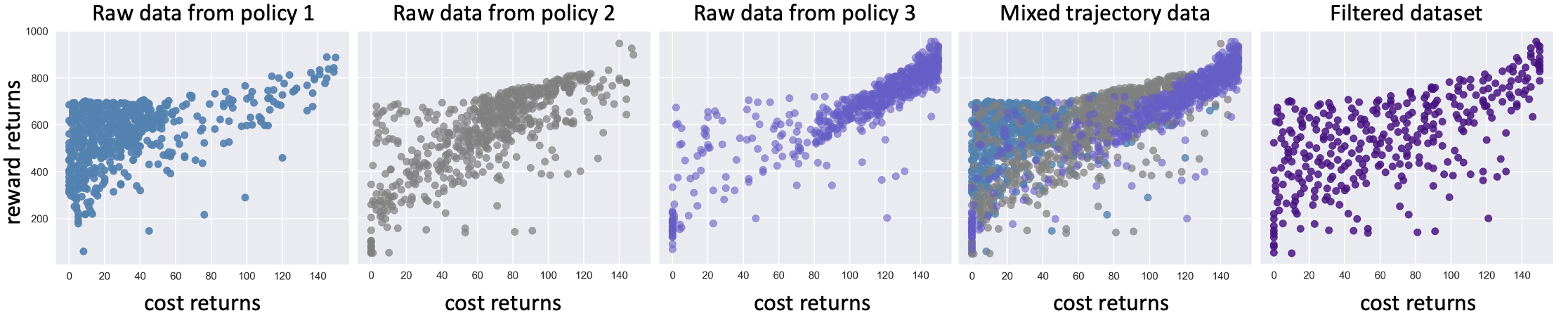}
    \caption{Illustration of the cost-reward plot and dataset collection process from a mixture of experts.}
    \label{fig:dataset-collect}
    \vspace{-2mm}
\end{figure}

Previous work suggests the cost-reward plot to assess and visualize the diversity of offline datasets with both reward and cost metrics, as it can significantly affect task complexity and training difficulty \cite{liu2023constrained}. 
Specifically, for each trajectory, we compute its total reward and total cost. Plotting these points on a two-dimensional plane where the x-axis represents the total cost and the y-axis represents the total reward, we obtain a scatter plot that characterizes the trade-offs between reward maximization and constraint satisfaction within the dataset. As Figure \ref{fig:dataset-collect} shows, the spread of points on this plot indicates the diversity within the dataset. 
Moreover, the plot's shape, specifically its reward frontiers regarding the cost, influences the task's complexity. 
Trajectories with high rewards but also high costs present tempting but risky opportunities for offline learners \cite{liu2022robustness, liu2023constrained}.
Therefore, the cost-reward plot serves as an intuitive tool for understanding the dataset's property, complexity, and diversity. 
This, in turn, aids in selecting appropriate datasets for benchmarking offline safe RL algorithms.

\vspace{-1mm}
\section{Datasets and Benchmarks}
\label{sec:method}
\vspace{-1mm}
\subsection{Dataset Collection}
\vspace{-1mm}
Our objective is to collect an array of high-quality datasets that span a spectrum of difficulty levels, thus enabling an unbiased evaluation of various algorithms' capabilities.
With this goal, we implement the \texttt{FSRL} library, containing advanced safe RL algorithms with carefully-tuned hyper-parameters, to generate a broad spectrum of datasets. The supported algorithms include the PID Lagrangian-based methods \cite{ray2019benchmarking, stooke2020responsive}, first-order method \cite{zhang2020first}, second-order method \cite{achiam2017constrained}, and variational-inference-based method \cite{liu2022constrained}.
Each task is trained by a suite of expert policies subjected to varying cost thresholds, thereby obtaining a pool of raw data that captures the intricacies of different task scenarios.

Subsequently, we apply a density filter across the cost-reward return space. This filter removes redundant trajectories that are highly concentrated within the same region, thereby maintaining greater diversity within the dataset.
The full procedure is visualized in Figure \ref{fig:dataset-collect}.
More details regarding the implemented expert algorithms, training tricks, and hyperparameters are available in the supplementary material.
Through this methodical approach to data collection, we strive to provide a rich and varied foundation for assessing the strengths and limitations of offline safe RL algorithms under a broad range of conditions.

\begin{figure}[b]
\vspace{-3mm}
    \centering    \includegraphics[width=0.99\linewidth]{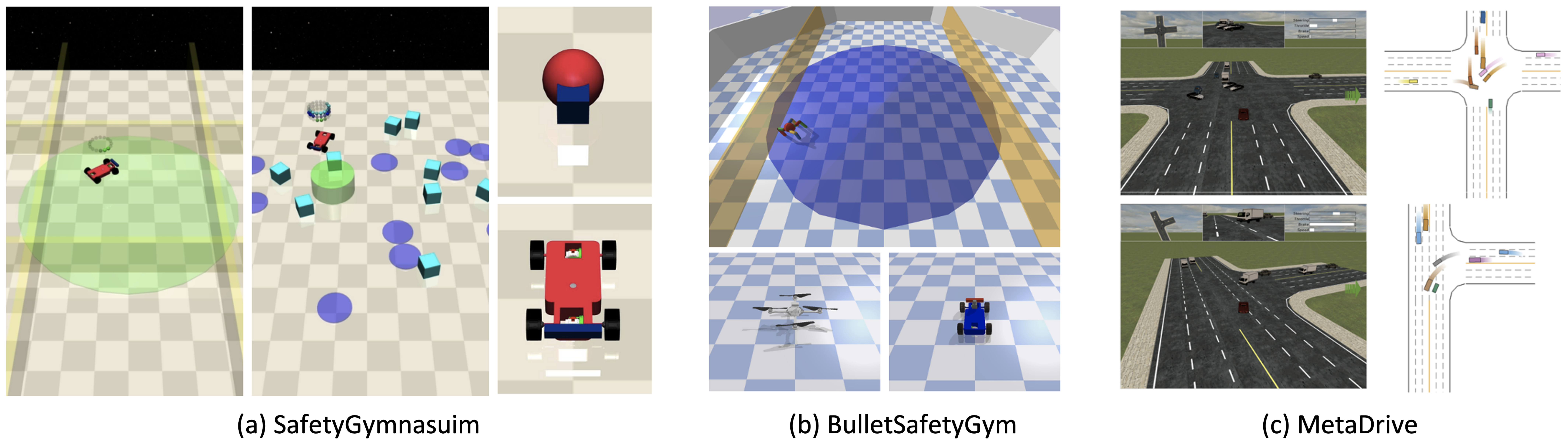}
    \vspace{-2mm}
    \caption{Visualization of the simulation environments and representative tasks.}
    \label{fig:task-visualization}
    \vspace{-3mm}
\end{figure}

\textbf{Simulation Environments and Tasks.} We gather datasets from three widely recognized safe RL environments: \textbf{1) SafetyGymnasium} \cite{ray2019benchmarking, ji2023omnisafe}, a collection of environments based on the Mujoco physics simulator, which offers a diverse range of tasks, with various safety constraints and challenges that can be adjusted to create different difficulty levels.
\textbf{2) BulletSafetyGym} \cite{gronauer2022bullet}, a suite of environments built on top of the PyBullet physics simulator, which is similar to SafetyGymnasium but with shorter horizons and more agents.
\textbf{3) MetaDrive} \cite{li2022metadrive}, a self-driving simulator based on the Panda3D game engine \cite{goslin2004panda3d}, which provides intricate road conditions and dynamic scenarios that closely emulate real-world driving situations, enabling the evaluation of safe RL algorithms in high-stakes, realistic applications.
Figure \ref{fig:task-visualization} visualize some representative tasks of these environments.

\begin{table}[ht]
\centering
\caption{\small Overview of the safe RL benchmarks and tasks for dataset collection} 
\label{tab:environment-overview}
\renewcommand{\arraystretch}{1.1}
\resizebox{0.9\linewidth}{!}{
\begin{tabular}{|c|c|c|c|c|c|c|}
\hline
Benchmarks                       & Backends                & Environments                                                         & Agents                                                                                & \begin{tabular}[c]{@{}c@{}}Difficulty\\ Levels\end{tabular} & \begin{tabular}[c]{@{}c@{}}Total\\ Tasks\end{tabular} & \begin{tabular}[c]{@{}c@{}}Dataset\\ Trajectories\end{tabular} \\ \hline
\multirow{2}{*}{SafetyGymnasium} & \multirow{2}{*}{Mujoco} & \begin{tabular}[c]{@{}c@{}}Goal, Button,\\ Push, Circle\end{tabular} & Point, Car                                                                            & 2                                                           & 16                                                    & 40310                                                          \\ \cline{3-7} 
                                 &                         & Velocity                                                             & \begin{tabular}[c]{@{}c@{}}Ant, HalfCheetah, Hopper,\\ Swimmer, Walker2d\end{tabular} & 1                                                           & 5                                                     & 11399                                                          \\ \hline
BulletSafetyGym                  & PyBullet                & Run, Circle                                                          & Ball, Car, Drone, Ant                                                                 & 1                                                           & 8                                                     & 14498                                                          \\ \hline
MetaDrive                        & Panda3D                 & Driving                                                              & Vehicle                                                                               & 3                                                           & 9                                                     & 9000                                                           \\ \hline
\end{tabular}
}
\end{table}

An overview of these environments and tasks is presented in Table \ref{tab:environment-overview}. We totally collect over 75000 diverse trajectories from 38 tasks.
A detailed breakdown of these datasets, including task names, trajectory size, observation space, and action space, can be found in the supplementary material.
By gathering datasets from these distinct environments, we ensure a well-rounded evaluation and benchmarking process that accurately reflects the capabilities of offline safe RL algorithms across a wide spectrum of tasks and complexities.

\subsection{Dataset Wrapper and Post-process Filters}
\label{sec:post-process}

We provide and maintain all the collected datasets via the \texttt{DSRL} package, which follows the same user-friendly API structure as D4RL \cite{fu2020d4rl}, facilitating the usage for researchers. The key distinction lies in the inclusion of a specialized \texttt{costs} entry in the datasets for indicating constraint violations. 

Apart from the access to the full datasets that are diverse over the cost-reward return space, we also provide a set of post-process filters to adjust the complexity and difficulty level of each dataset, aiming to achieve a comprehensive evaluation for different perspectives, as we will introduce in section \ref{sec:evaluating-offline-safe-rl}.
This involves changing data density, discarding data within specific cost-reward ranges, and increasing outlier trajectories. Here is a detailed description of these filters:

\begin{wrapfigure}{r}{0.57\textwidth}
  \begin{center}
  \vspace{-7mm}
  \includegraphics[width=0.57\textwidth]{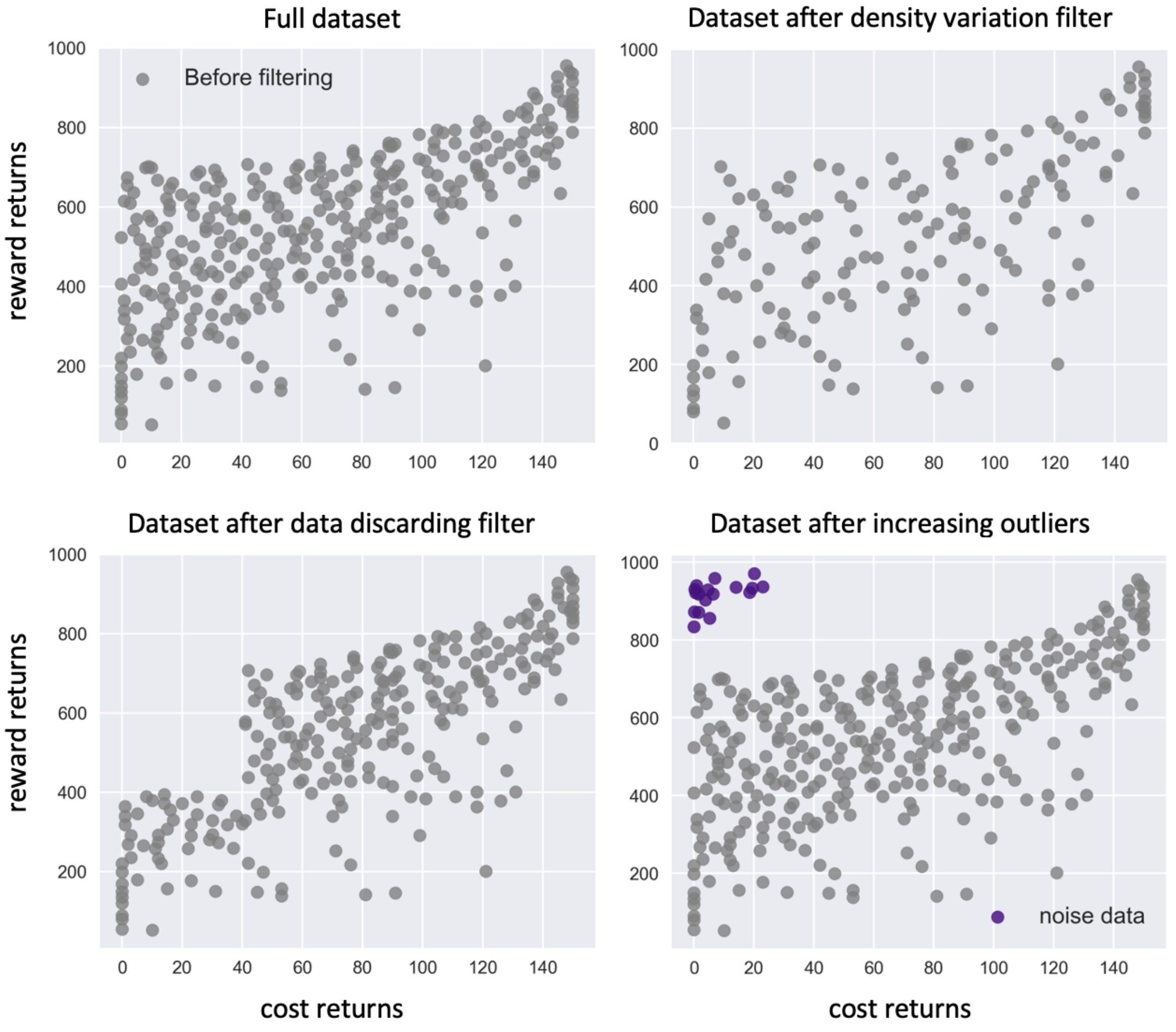}
  \end{center}
  \vspace{-4mm}
  \caption{\small Illustration of the post-process filters.}
  \label{fig:dataset-filters}
  \vspace{-3mm}
\end{wrapfigure}

\textbf{Filter for Data Density Variation}: This filter aims to create datasets with varying data densities. These variations will help evaluate the algorithms' ability to perform under different levels of data availability and assess their generalization capabilities.

\textbf{Filter for Partial Data Discarding}: 
This filter operates by selectively removing trajectories within defined return ranges, represented as $[(r_{\text{min}}, c_{\text{min}}), (r_{\text{max}}, c_{\text{max}})]$. Typically, by discarding trajectories with low costs and high rewards, we alter the reward distribution, mirroring scenarios where data is collected by either overly conservative or excessively risky policies. Another usage scenario is removing data within specific cost ranges, which can assess an algorithm's ability to manage unseen safety thresholds and learn from sub-optimal data. 

\textbf{Filter for Noise Level Manipulation}:
This filter introduces different degrees of outliers to the dataset trajectories. Specifically, we select a subset of high-cost trajectories and relabel their cost returns as low, simultaneously increasing their reward returns. This process simulates outlier trajectories that could potentially mislead learning algorithms by promoting hazardous behaviors, mimicking scenarios where abnormality arises during data collection. With this filter in place, we assess the algorithm's capacity to manage noisy data and its tolerance towards outliers.

These filters function deterministically, ensuring the resulting datasets remain consistent given a fixed parameter, thereby maintaining fair comparison between algorithms. 
In addition, they offer compositional flexibility when applied to the entire dataset, permitting customization of learning problem difficulty levels. As a result, our benchmark offers hundreds of unique datasets, each embodying different challenge levels, allowing us to efficiently evaluate various aspects of safe learning algorithms across a diverse range of complexity, as we will introduce in the next section.
We also present examples of using these filters in the experiments and supplementary material. 

\subsection{Evaluating Offline Safe RL Algorithms}
\label{sec:evaluating-offline-safe-rl}

Given the distinctive problem setting, objectives, and applications of offline safe RL compared to standard RL, we reevaluate the methodology for comparing offline safe RL approaches. Accordingly, we propose a tiered, four-level evaluation perspective, organized in order of importance, to assess safe offline reinforcement learning algorithms more accurately.

\noindent \textbf{Safety Compliance}: 
This primary requirement evaluates an agent's adherence to specific safety rules.
The learning agent must not violate any safety constraints while striving to maximize rewards. 

\textbf{Reward-seeking}: Alongside safety, we assess the agent's ability to maximize rewards within the safety set. We want to prevent the agent from being overly conservative and achieve a balance between safety and reward-seeking behavior.

While these initial levels focus on fundamental requirements, the remaining levels target more advanced attributes that add value to offline safe RL algorithms.

\noindent \textbf{Generalization}: A generalizable safe learning algorithm should be capable of learning from sub-optimal data and be robust to unseen samples upon deployment, as obtaining datasets that comprehensively cover every possible scenario is impractical. 
This perspective measures the agent's adaptability to previously unseen training conditions and requirements.

\noindent \textbf{Outlier Sensitivity}: 
This attribute tests the algorithm's resilience against outliers in the datasets. The algorithm's performance against outlier trajectories with abnormally high rewards and low costs is assessed, ensuring it can effectively handle noisy or imperfect data without compromising safety.

We present these evaluation perspectives as a guiding resource for future researchers and practitioners.
While our benchmark primarily centers on the first two levels of evaluation, we supply an array of filters to post-process the datasets. These filters aid in assessing the generalization capabilities and outlier sensitivity of learning algorithms, as we presented in the previous section.

\subsection{Offline Safe RL Benchmarks}
\label{sec:osrl-benchmark}

We implemented a broad range of existing offline safe RL methods in our benchmark, all organized under the \texttt{OSRL} package, whose framework design is mainly inspired by the user-friendly CORL \cite{tarasov2022corl} library.
A detailed summary of these methods is provided in Table \ref{tab:osrl-algorithms}. The \textit{Type} row categorizes the learning algorithm (such as Q-learning or imitation learning), while the \textit{Base Method} row indicates the corresponding offline learning method, excluding safety constraints considerations. As far as we know, these implemented algorithms represent the majority of the offline safe learning categories currently available in the literature.

It's worth highlighting that the Lagrangian-based methods align with the expert safe RL policy implementation in our \texttt{FSRL} package, which utilizes adaptive PID-based Lagrangian multipliers to penalize constraint violations \cite{stooke2020responsive}. This approach can be effortlessly extended to other existing Q-learning-based offline RL methods.
In addition, we've prioritized coherence and user-friendliness in the API structure of these implementations, aiming to provide a valuable resource that can contribute to and further the development of the offline safe RL community.



\begin{table}[ht]
\centering
\caption{\small Implemented offline safe learning algorithms and their base methods.} 
\label{tab:osrl-algorithms}
\renewcommand{\arraystretch}{1.1}
\resizebox{1.\linewidth}{!}{
\begin{tabular}{|c|c|c|c|ccc|}
\hline
Type        & Sequential Modeling & Imitation Learning & \begin{tabular}[c]{@{}c@{}}Distribution Correction\\ Estimation\end{tabular} & \multicolumn{3}{c|}{Q-learning}                                                    \\ \hline
Algorithm   & CDT \cite{liu2023constrained}                & BC-\{Safe, All\}  \cite{liu2023constrained, xu2022constraints}          & COptiDICE   \cite{lee2022coptidice}                                                                 & \multicolumn{1}{c|}{CPQ \cite{xu2022constraints}} & \multicolumn{1}{c|}{BCQ-Lag \cite{xu2022constraints}}         & BEAR-Lag \cite{xu2022constraints}        \\ \hline
Base Method & \begin{tabular}[c]{@{}c@{}}Decision \\ Transformer \cite{chen2021decision}\end{tabular}                  & Behavior Cloning                 & OptiDICE  \cite{lee2021optidice}                                                                  & \multicolumn{1}{c|}{BCQ \cite{fujimoto2019off}} & \multicolumn{1}{c|}{ \begin{tabular}[c]{@{}c@{}} BCQ \cite{fujimoto2019off}, \\ Lagrangian \cite{stooke2020responsive} \end{tabular}  } & \begin{tabular}[c]{@{}c@{}}BEAR \cite{kumar2019stabilizing}, \\ Lagrangian \cite{stooke2020responsive}\end{tabular}   \\ \hline
\end{tabular}
}
\end{table}

\subsection{Evaluation Metrics}
\label{sec:evaluation-metric}
We adopt the normalized reward return and the normalized cost return as the comparison metrics \cite{fu2020d4rl, liu2023constrained}.
Denote $r_{\text{max}}(\Mcal)$ and $r_{\text{min}}(\Mcal)$ as the maximum empirical reward return and the minimum empirical reward return for task $\Mcal$. The normalized reward is computed by:
$$
R_{\text{normalized}} = \frac{R_\pi - r_{\text{min}}(\Mcal)}{r_{\text{max}}(\Mcal)-r_{\text{min}}(\Mcal)} \times 100,
$$
where $R_\pi$ denotes the evaluated reward return of policy $\pi$. Note that we use a constant maximum and minimum values for a safe RL task rather than a dataset. This is because the post-process filters may modify the dataset to create different difficulty levels.

The normalized cost is defined differently from the reward to better distinguish the results. It is computed by the ratio between the evaluated cost return $C_\pi$ and the target threshold $\kappa$:
$$
C_{\text{normalized}} = \frac{C_\pi + \epsilon}{\kappa + \epsilon},
$$
where $\epsilon$ is a positive number to ensure numerical stability if the threshold $\kappa=0$. Note that the cost return and the threshold are always non-negative.
Without otherwise statements, we will abbreviate ``normalized cost return" as ``cost" and ``normalized reward return" as ``reward" for simplicity.

\vspace{-1mm}
\section{Experiments and Analysis}
\vspace{-1mm}
\label{sec:experiment}

\textbf{Experiment Settings.}
Contrary to earlier safe RL studies, which tested agents under a single threshold, we adopt a \textbf{Constraint Variation Evaluation} to assess algorithm versatility.
By training agents with varying safety constraint requirements, we can evaluate an algorithm's adaptability to a diverse range of safety conditions. 
This is similar to the Average Precision or the Area Under the Curve metrics in the literature \cite{davis2006relationship}.
More specifically, each algorithm is evaluated on each dataset using three distinct target thresholds and with three random seeds.
We then compute the average of the normalized reward and cost to characterize the performance on varying safety conditions better.

\begin{table}[ht]
\centering
\caption{\small Evaluation results of the normalized reward and cost. The cost threshold is 1.
The $\uparrow$ symbol denotes that the higher reward, the better. The $\downarrow$ symbol denotes that the lower cost (up to threshold 1), the better. 
Each value is averaged over 3 distinct cost thresholds, 20 evaluation episodes, and 3 random seeds.
\textbf{Bold}: Safe agents whose normalized cost is smaller than 1. 
{\color[HTML]{656565} Gray}: Unsafe agents.
{\color[HTML]{0000FF} \textbf{Blue}}: Safe agent with the highest reward.} 
\label{tab:main-result}
\renewcommand{\arraystretch}{1.1}
\resizebox{1.\linewidth}{!}{
\begin{tabular}{|c|cc|cc|cc|cc|cc|cc|cc|}
\hline
                                                                  & \multicolumn{2}{c|}{BC-All}                                & \multicolumn{2}{c|}{BC-Safe}                                                 & \multicolumn{2}{c|}{CDT}                                                    & \multicolumn{2}{c|}{BCQ-Lag}                                                & \multicolumn{2}{c|}{BEAR-Lag}                              & \multicolumn{2}{c|}{CPQ}                                    & \multicolumn{2}{c|}{COptiDICE}                                              \\ \cline{2-15} 
\multirow{-2}{*}{Task}                                            & reward $\uparrow$                       & cost $\downarrow$                        & reward $\uparrow$                                & cost $\downarrow$                                 & reward $\uparrow$                               & cost $\downarrow$                                 & reward $\uparrow$                               & cost $\downarrow$                                 & reward $\uparrow$                       & cost $\downarrow$                        & reward $\uparrow$                       & cost $\downarrow$                         & reward $\uparrow$                               & cost $\downarrow$                                 \\ \hline
CarButton1                                                        & {\color[HTML]{656565} 0.03}  & {\color[HTML]{656565} 1.38} & {\color[HTML]{0000FF} \textbf{0.07}}  & {\color[HTML]{0000FF} \textbf{0.85}} & {\color[HTML]{656565} 0.21}          & {\color[HTML]{656565} 1.6}           & {\color[HTML]{656565} 0.04}          & {\color[HTML]{656565} 1.63}          & {\color[HTML]{656565} 0.18}  & {\color[HTML]{656565} 2.72} & {\color[HTML]{656565} 0.42}  & {\color[HTML]{656565} 9.66}  & {\color[HTML]{656565} -0.08}         & {\color[HTML]{656565} 1.68}          \\
CarButton2                                                        & {\color[HTML]{656565} -0.13} & {\color[HTML]{656565} 1.24} & {\color[HTML]{0000FF} \textbf{-0.01}} & {\color[HTML]{0000FF} \textbf{0.63}} & {\color[HTML]{656565} 0.13}          & {\color[HTML]{656565} 1.58}          & {\color[HTML]{656565} 0.06}          & {\color[HTML]{656565} 2.13}          & {\color[HTML]{656565} -0.01} & {\color[HTML]{656565} 2.29} & {\color[HTML]{656565} 0.37}  & {\color[HTML]{656565} 12.51} & {\color[HTML]{656565} -0.07}         & {\color[HTML]{656565} 1.59}          \\
CarGoal1                                                          & \textbf{0.39}                & \textbf{0.33}               & \textbf{0.24}                         & \textbf{0.28}                        & {\color[HTML]{656565} 0.66}          & {\color[HTML]{656565} 1.21}          & {\color[HTML]{0000FF} \textbf{0.47}} & {\color[HTML]{0000FF} \textbf{0.78}} & {\color[HTML]{656565} 0.61}  & {\color[HTML]{656565} 1.13} & {\color[HTML]{656565} 0.79}  & {\color[HTML]{656565} 1.42}  & \textbf{0.35}                        & \textbf{0.54}                        \\
CarGoal2                                                          & {\color[HTML]{656565} 0.23}  & {\color[HTML]{656565} 1.05} & \textbf{0.14}                         & \textbf{0.51}                        & {\color[HTML]{656565} 0.48}          & {\color[HTML]{656565} 1.25}          & {\color[HTML]{656565} 0.3}           & {\color[HTML]{656565} 1.44}          & {\color[HTML]{656565} 0.28}  & {\color[HTML]{656565} 1.01} & {\color[HTML]{656565} 0.65}  & {\color[HTML]{656565} 3.75}  & {\color[HTML]{0000FF} \textbf{0.25}} & {\color[HTML]{0000FF} \textbf{0.91}} \\
CarPush1                                                          & \textbf{0.22}                & \textbf{0.36}               & \textbf{0.14}                         & \textbf{0.33}                        & {\color[HTML]{0000FF} \textbf{0.31}} & {\color[HTML]{0000FF} \textbf{0.4}}  & \textbf{0.23}                        & \textbf{0.43}                        & \textbf{0.21}                & \textbf{0.54}               & \textbf{-0.03}               & \textbf{0.95}                & \textbf{0.23}                        & \textbf{0.5}                         \\
CarPush2                                                          & {\color[HTML]{0000FF} \textbf{0.14}}                & {\color[HTML]{0000FF} \textbf{0.9}}                & \textbf{0.05}                         & \textbf{0.45}                        & {\color[HTML]{656565} 0.19}          & {\color[HTML]{656565} 1.3}           & {\color[HTML]{656565} 0.15}          & {\color[HTML]{656565} 1.38}          & {\color[HTML]{656565} 0.1}   & {\color[HTML]{656565} 1.2}  & {\color[HTML]{656565} 0.24}  & {\color[HTML]{656565} 4.25}  & {\color[HTML]{656565} 0.09}          & {\color[HTML]{656565} 1.07}          \\
PointButton1                                                      & {\color[HTML]{656565} 0.1}   & {\color[HTML]{656565} 1.05} & {\color[HTML]{0000FF} \textbf{0.06}}                         & {\color[HTML]{0000FF} \textbf{0.52}}                        & {\color[HTML]{656565} 0.53}          & {\color[HTML]{656565} 1.68}          & {\color[HTML]{656565} 0.24}          & {\color[HTML]{656565} 1.73}          & {\color[HTML]{656565} 0.2}   & {\color[HTML]{656565} 1.6}  & {\color[HTML]{656565} 0.69}  & {\color[HTML]{656565} 3.2}   & {\color[HTML]{656565} 0.13}          & {\color[HTML]{656565} 1.35}          \\
PointButton2                                                      & {\color[HTML]{656565} 0.27}  & {\color[HTML]{656565} 2.02} & {\color[HTML]{656565} 0.16}           & {\color[HTML]{656565} 1.1}           & {\color[HTML]{656565} 0.46}          & {\color[HTML]{656565} 1.57}          & {\color[HTML]{656565} 0.4}           & {\color[HTML]{656565} 2.66}          & {\color[HTML]{656565} 0.43}  & {\color[HTML]{656565} 2.47} & {\color[HTML]{656565} 0.58}  & {\color[HTML]{656565} 4.3}   & {\color[HTML]{656565} 0.15}          & {\color[HTML]{656565} 1.51}          \\
PointGoal1                                                        & \textbf{0.65}                & \textbf{0.95}               & \textbf{0.43}                         & \textbf{0.54}                        & {\color[HTML]{656565} 0.69}          & {\color[HTML]{656565} 1.12}          & {\color[HTML]{0000FF} \textbf{0.71}} & {\color[HTML]{0000FF} \textbf{0.98}} & 0.74                         & 1.18                        & \textbf{0.57}                & \textbf{0.35}                & {\color[HTML]{656565} 0.49}          & {\color[HTML]{656565} 1.66}          \\
PointGoal2                                                        & {\color[HTML]{656565} 0.54}  & {\color[HTML]{656565} 1.97} & {\color[HTML]{0000FF} \textbf{0.29}}  & {\color[HTML]{0000FF} \textbf{0.78}} & {\color[HTML]{656565} 0.59}          & {\color[HTML]{656565} 1.34}          & {\color[HTML]{656565} 0.67}          & {\color[HTML]{656565} 3.18}          & {\color[HTML]{656565} 0.67}  & {\color[HTML]{656565} 3.11} & {\color[HTML]{656565} 0.4}   & {\color[HTML]{656565} 1.31}  & {\color[HTML]{656565} 0.38}          & {\color[HTML]{656565} 1.92}          \\
PointPush1                                                        & \textbf{0.19}                & \textbf{0.61}               & \textbf{0.13}                         & \textbf{0.43}                        & \textbf{0.24}                        & \textbf{0.48}                        & {\color[HTML]{0000FF} \textbf{0.33}} & {\color[HTML]{0000FF} \textbf{0.86}} & \textbf{0.22}                & \textbf{0.79}               & \textbf{0.2}                 & \textbf{0.83}                & \textbf{0.13}                        & \textbf{0.83}                        \\
PointPush2                                                        & \textbf{0.18}                & \textbf{0.91}               & \textbf{0.11}                         & \textbf{0.8}                         & \textbf{0.21}                        & \textbf{0.65}                        & {\color[HTML]{0000FF} \textbf{0.23}} & {\color[HTML]{0000FF} \textbf{0.99}} & \textbf{0.16}                & \textbf{0.89}               & 0.11                         & 1.04                         & 0.02                                 & 1.18                                 \\
HalfCheetahVel                                               & {\color[HTML]{656565} 0.97}  & {\color[HTML]{656565} 13.1} & \textbf{0.88}                         & \textbf{0.54}                        & {\color[HTML]{0000FF} \textbf{1.0}}  & {\color[HTML]{0000FF} \textbf{0.01}} & {\color[HTML]{656565} 1.05}          & {\color[HTML]{656565} 18.21}         & {\color[HTML]{656565} 0.98}  & {\color[HTML]{656565} 6.58} & \textbf{0.29}                & \textbf{0.74}                & \textbf{0.65}                        & \textbf{0.0}                         \\
SwimmerVel                                                   & {\color[HTML]{656565} 0.49}  & {\color[HTML]{656565} 4.72} & {\color[HTML]{656565} 0.51}           & {\color[HTML]{656565} 1.07}          & {\color[HTML]{0000FF} \textbf{0.66}} & {\color[HTML]{0000FF} \textbf{0.96}} & {\color[HTML]{656565} 0.48}          & {\color[HTML]{656565} 6.58}          & {\color[HTML]{656565} 0.3}   & {\color[HTML]{656565} 2.33} & {\color[HTML]{656565} 0.13}  & {\color[HTML]{656565} 2.66}  & {\color[HTML]{656565} 0.63}          & {\color[HTML]{656565} 7.58}          \\
Walker2dVel                                                  & {\color[HTML]{656565} 0.79}  & {\color[HTML]{656565} 3.88} & {\color[HTML]{0000FF} \textbf{0.79}}  & {\color[HTML]{0000FF} \textbf{0.04}} & \textbf{0.78}                        & \textbf{0.06}                        & \textbf{0.79}                        & \textbf{0.17}                        & {\color[HTML]{656565} 0.86}  & {\color[HTML]{656565} 3.1}  & \textbf{0.04}                & \textbf{0.21}                & \textbf{0.12}                        & \textbf{0.74}                        \\ \hline
\begin{tabular}[c]{@{}c@{}} \textbf{SafetyGym}\\ \textbf{Average}\end{tabular} & {\color[HTML]{656565} 0.34}  & {\color[HTML]{656565} 2.30} & {\color[HTML]{0000FF} \textbf{0.27}}   & {\color[HTML]{0000FF} \textbf{0.60}} & {\color[HTML]{656565} 0.48}          & {\color[HTML]{656565} 1.01}          & {\color[HTML]{656565} 0.41}          & {\color[HTML]{656565} 2.88}          & {\color[HTML]{656565} 0.40}  & {\color[HTML]{656565} 2.06} & {\color[HTML]{656565} 0.36}  & {\color[HTML]{656565} 3.15}  & {\color[HTML]{656565} 0.23}          & {\color[HTML]{656565} 1.54}          \\ \hline
AntCircle                                                         & {\color[HTML]{656565} 0.58}  & {\color[HTML]{656565} 4.9}  & {\color[HTML]{0000FF} \textbf{0.4}}                          & {\color[HTML]{0000FF} \textbf{0.96}}                        & 0.54                                 & 1.78                                 & {\color[HTML]{656565} 0.58}          & {\color[HTML]{656565} 2.87}          & {\color[HTML]{656565} 0.65}  & {\color[HTML]{656565} 5.48} & \textbf{0.0}                 & \textbf{0.0}                 & {\color[HTML]{656565} 0.17}          & {\color[HTML]{656565} 5.04}          \\
AntRun                                                            & {\color[HTML]{656565} 0.72}  & {\color[HTML]{656565} 2.93} & 0.65                                  & 1.09                                 & {\color[HTML]{0000FF} \textbf{0.72}} & {\color[HTML]{0000FF} \textbf{0.91}} & {\color[HTML]{656565} 0.76}          & {\color[HTML]{656565} 5.11}          & \textbf{0.15}                & \textbf{0.73}               & \textbf{0.03}                & \textbf{0.02}                & \textbf{0.61}                        & \textbf{0.94}                        \\
CarCircle                                                         & {\color[HTML]{656565} 0.58}  & {\color[HTML]{656565} 3.74} & \textbf{0.5}                          & \textbf{0.84}                        & {\color[HTML]{0000FF} \textbf{0.75}} & {\color[HTML]{0000FF} \textbf{0.95}} & {\color[HTML]{656565} 0.63}          & {\color[HTML]{656565} 1.89}          & {\color[HTML]{656565} 0.74}  & {\color[HTML]{656565} 2.18} & \textbf{0.71}                & \textbf{0.33}                & {\color[HTML]{656565} 0.49}          & {\color[HTML]{656565} 3.14}          \\
DroneCircle                                                       & {\color[HTML]{656565} 0.72}  & {\color[HTML]{656565} 3.03} & \textbf{0.56}                         & \textbf{0.57}                        & {\color[HTML]{0000FF} \textbf{0.63}} & {\color[HTML]{0000FF} \textbf{0.98}} & {\color[HTML]{656565} 0.8}           & {\color[HTML]{656565} 3.07}          & {\color[HTML]{656565} 0.78}  & {\color[HTML]{656565} 3.68} & {\color[HTML]{656565} -0.22} & {\color[HTML]{656565} 1.28}  & {\color[HTML]{656565} 0.26}          & {\color[HTML]{656565} 1.02}          \\
DroneRun                                                          & {\color[HTML]{656565} 0.24}  & {\color[HTML]{656565} 2.13} & \textbf{0.28}                         & \textbf{0.74}                        & {\color[HTML]{0000FF} \textbf{0.63}} & {\color[HTML]{0000FF} \textbf{0.79}} & {\color[HTML]{656565} 0.72}          & {\color[HTML]{656565} 5.54}          & {\color[HTML]{656565} 0.42}  & {\color[HTML]{656565} 2.47} & {\color[HTML]{656565} 0.33}  & {\color[HTML]{656565} 3.52}  & {\color[HTML]{656565} 0.67}          & {\color[HTML]{656565} 4.15}          \\ \hline
\begin{tabular}[c]{@{}c@{}}\textbf{BulletGym}\\ \textbf{Average}\end{tabular} & {\color[HTML]{656565} 0.57}  & {\color[HTML]{656565} 3.35} & {\color[HTML]{0000FF} \textbf{0.48}}  & {\color[HTML]{0000FF} \textbf{0.84}} & 0.66                                 & 1.08                                 & {\color[HTML]{656565} 0.7}           & {\color[HTML]{656565} 3.7}           & {\color[HTML]{656565} 0.55}  & {\color[HTML]{656565} 2.91} & {\color[HTML]{656565} 0.17}  & {\color[HTML]{656565} 1.03}  & {\color[HTML]{656565} 0.44}          & {\color[HTML]{656565} 2.86}          \\ \hline
easydense                                                         & {\color[HTML]{656565} 0.27}  & {\color[HTML]{656565} 1.94} & \textbf{0.11}                         & \textbf{0.14}                        & {\color[HTML]{0000FF} \textbf{0.32}} & {\color[HTML]{0000FF} \textbf{0.62}} & \textbf{0.26}                        & \textbf{0.47}                        & \textbf{0.02}                & \textbf{0.41}               & \textbf{-0.06}               & \textbf{0.03}                & {\color[HTML]{656565} 0.5}           & {\color[HTML]{656565} 2.54}          \\
mediummean                                                        & {\color[HTML]{656565} 0.77}  & {\color[HTML]{656565} 2.53} & \textbf{0.31}                         & \textbf{0.21}                        & {\color[HTML]{0000FF} \textbf{0.45}} & {\color[HTML]{0000FF} \textbf{0.75}} & {\color[HTML]{656565} 0.78}          & {\color[HTML]{656565} 1.53}          & \textbf{-0.0}                & \textbf{0.34}               & \textbf{-0.08}               & \textbf{0.05}                & {\color[HTML]{656565} 0.76}          & {\color[HTML]{656565} 2.05}          \\
hardsparse                                                        & {\color[HTML]{656565} 0.42}  & {\color[HTML]{656565} 1.8}  & {\color[HTML]{656565} 0.17}           & {\color[HTML]{656565} 3.25}          & {\color[HTML]{0000FF} \textbf{0.25}} & {\color[HTML]{0000FF} \textbf{0.41}} & {\color[HTML]{656565} 0.5}           & {\color[HTML]{656565} 1.02}          & \textbf{0.01}                & \textbf{0.16}               & \textbf{-0.05}               & \textbf{0.06}                & {\color[HTML]{656565} 0.37}          & {\color[HTML]{656565} 2.05}          \\ \hline
\begin{tabular}[c]{@{}c@{}}\textbf{MetaDrive}\\ \textbf{Average}\end{tabular}       & {\color[HTML]{656565} 0.49}  & {\color[HTML]{656565} 2.09} & {\color[HTML]{656565} 0.2}            & {\color[HTML]{656565} 1.2}           & {\color[HTML]{0000FF} \textbf{0.34}} & {\color[HTML]{0000FF} \textbf{0.59}} & {\color[HTML]{656565} 0.51}          & {\color[HTML]{656565} 1.01}          & \textbf{0.01}                & \textbf{0.3}                & \textbf{-0.06}               & \textbf{0.05}                & {\color[HTML]{656565} 0.55}          & {\color[HTML]{656565} 2.22}          \\ \hline
\end{tabular}
}
\end{table}
The evaluation results from representative datasets for each environment are presented in Table \ref{tab:main-result}. Here, \textbf{BC-All} refers to behavior cloning trained with all datasets, while \textbf{BC-Safe} refers to behavior cloning trained exclusively with safe trajectories that satisfy the constraints. The complete results on all datasets and detailed hyperparameters can be found in the supplementary material.

\textbf{Main Results Analysis}. The performance of tested algorithms provides valuable insights into the challenges of offline safe learning. 
\textbf{BC-All} and \textbf{BC-Safe}, focusing on imitating policies rather than estimating Q values, exhibit stark differences: BC-All achieves higher rewards but fails on safety constraints; BC-Safe, fed with only safe trajectories, satisfies most safety requirements, although with conservative performances and lower rewards. This comparison underlines the essential trade-offs between safety and utility in offline safe RL, largely dictated by the training dataset used.

\textbf{CDT}, through its advanced architecture and effective data utilization, offers a more balanced performance. Despite struggling with complex tasks in high-stochasticity environments, CDT generally yields higher rewards while maintaining safety, outperforming BC-Safe in most tasks. 

Contrarily, all Q-learning-based algorithms, including \textbf{BCQ-Lag}, \textbf{BEAR-Lag}, and \textbf{CPQ}, as well as \textbf{COptiDICE}, display performance inconsistencies, vacillating between excessive conservatism and riskiness. CPQ, for example, obtains high rewards at significant safety compromise in \texttt{Button} tasks, while achieving almost zero cost with low rewards in \texttt{MetaDrive} tasks. 

These inconsistencies expose the key challenge for Q-learning-based approaches in offline safe RL: accurately estimating the safety performance of trained policies. In standard offline RL, minor biases in Q estimation rarely impact overall performance. However, the safety thresholds introduce new dynamics. Under-estimating cost Q values could result in negligible safety penalties, causing overly risky policies, while overestimations can lead to overly conservative behaviors. 

To tackle this challenge, future research could focus on developing techniques for precise safety performance estimation in offline environments. This is particularly crucial for the application and evolution of Q-learning-based approaches in offline safe RL.

\begin{figure}[h]
\vspace{-1mm}
    \centering    \includegraphics[width=0.99\linewidth]{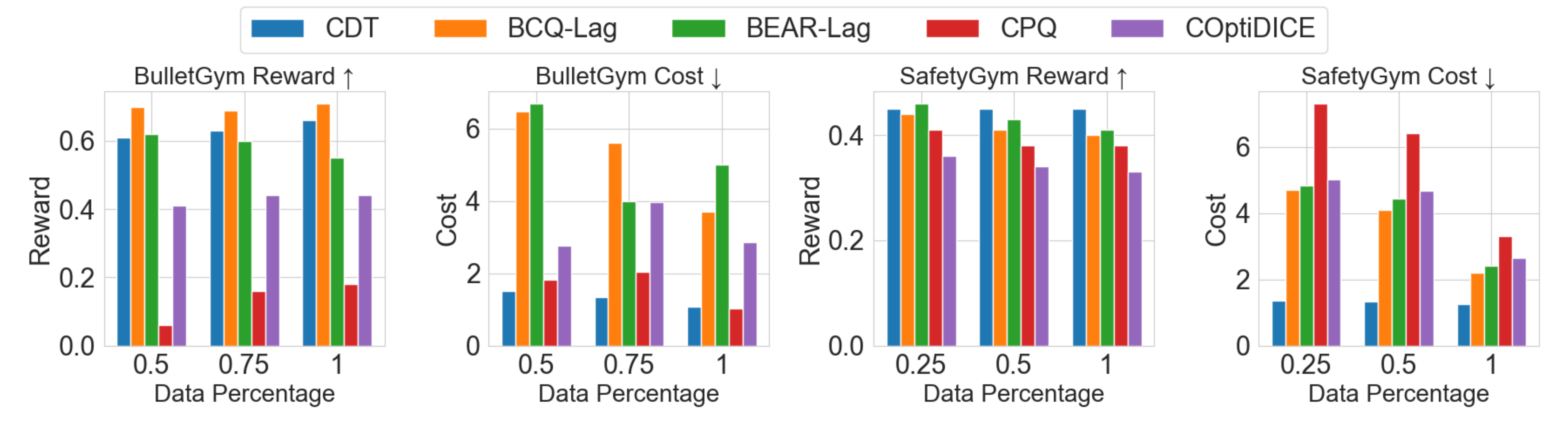}
    \vspace{-1mm}
    \caption{\small Average performance with different percentage of dataset trajectories.}
    \label{fig:bullet-density}
    \vspace{-1mm}
\end{figure}

\textbf{Post-process Filters Experiments.}
Our experiments further delve into a variety of evaluation criteria, as outlined in Sec. \ref{sec:evaluating-offline-safe-rl}, through the implementation of different data manipulation filters, notably density and noise level manipulation filters. 
The results are averaged among the corresponding Bullet-Gym and Safety-Gym suites of tasks. More details are available in the supplementary material.

As displayed in Fig. \ref{fig:bullet-density}, applying a density filter to vary dataset sizes reveals a trend: most algorithms exhibit a decrease in cost values as more data is employed. This finding underscores the pivotal role of dataset size in influencing algorithmic performance. Interestingly, the safety performance of BCQ-Lag and CPQ are noticeably influenced by data sizes, suggesting that certain algorithms may be more susceptible to data density. In contrast, CDT showcases robustness against sparse data, indicating its potential utility in environments where data collection may be challenging.

\begin{figure}[h]
\vspace{-2mm}
    \centering    \includegraphics[width=0.99\linewidth]{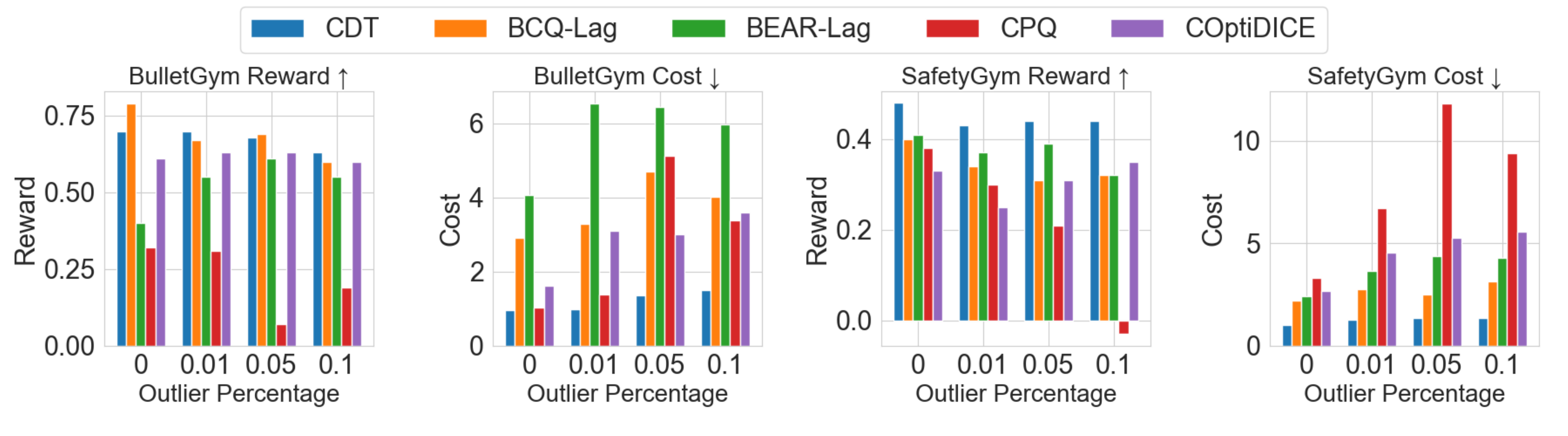}
    \vspace{-2mm}
    \caption{\small Average performance with different percentage of outlier trajectories.}
    \label{fig:noise}
    \vspace{-2mm}
\end{figure}

Fig. \ref{fig:noise} illustrates the performance impact of the noise data filter, which increases the percentage of outlier trajectories. Several algorithms, such as CPQ in SafetyGym tasks, exhibit drastic performance degradation, manifested as a substantial reward reduction and cost increase. This phenomenon underscores the critical importance of outlier sensitivity in the evaluation of offline safe RL algorithms. Excessive sensitivity to outliers may introduce instability during the learning process, leading to suboptimal safety and performance outcomes.

These experiments offer insights into the robustness and generalization capability of offline safe RL algorithms in the face of dataset variations. 
From a broader perspective, these findings illuminate the intricate dynamics and potential difficulties inherent in offline safe RL. The challenge lies in maintaining a delicate balance between reward optimization and safety assurance. Algorithms' performance on this front further underscores the complexity of the datasets used, emphasizing the need for more research into handling these challenges. Through our benchmarks, we hope to foster a deeper understanding of offline safe RL and to accelerate its real-world applications.

\vspace{-1mm}
\section{Challenges and Future Directions}
\label{sec:limitation}
\vspace{-1mm}
While our benchmark study has sought to cover a comprehensive range of factors in offline safe reinforcement learning, there are also limitations to be acknowledged.
For example, our study uses only three prevalent safe RL environments. While they provide a broad context for algorithm evaluation, we encourage the community to contribute more diverse and ideally real-world datasets to this field.
There are also many challenges in RL safety that remain untouched.
Below we outline some of these key areas, open problems, and potential future directions.


1. \textbf{Performance Metrics and Benchmarks}: While we have provided an evaluation framework, there is still room for more sophisticated performance metrics. These could account for various real-world factors, such as environmental changes \cite{chen2021context}, adversarial attacks \cite{liu2022robustness}, and more.

2. \textbf{Interpretability and Safety Certification}: As these algorithms become more complex, the need for explainability and theoretical safety guarantees become more pressing \cite{amodei2016concrete, verma2018programmatically, luo2021learning}. Ensuring that the actions and decisions of these algorithms can be understood by humans and can be certified to be safe will be crucial for their wider acceptance and adoption.

3. \textbf{Data Efficiency}: Offline safe RL algorithms are trained from datasets, making data efficiency a critical aspect \cite{schwarzer2021pretraining}. Future research could focus on improving data utilization, possibly through advances in sample-efficient learning techniques or data augmentation strategies \cite{sinha2022s4rl, as2022constrained}.

4. \textbf{Few-shot Online Adaptation}: Offline pretraining plus online adaptation is becoming a popular training paradigm in wide domains \cite{kumar2022pre, radosavovic2023real}. Therefore, the ability of offline safe learning algorithms to adapt to new environments safely with few shot samples is an area ripe for research \cite{ zhu2020transfer}.

5. \textbf{Versatility}: The capacity of an algorithm to adapt to varying safety constraints without the need for substantial re-tuning or re-training is pivotal. Currently, only sequential-modeling-based methods \cite{liu2023constrained, zhang2023saformer} can effectively achieve this, but there is substantial room for improvement. 

6. \textbf{Ethics and Fairness}: As offline learning is increasingly deployed in sensitive areas such as healthcare, considerations around ethics and fairness become particularly important \cite{jabbari2017fairness, thomas2019preventing, deng2022reinforcement}. Future work could focus on integrating these factors into the offline safe RL framework.

Addressing these challenges will help drive the field forward, pushing the boundaries of what offline safe RL can achieve. We aspire for the discussions and resources furnished in this work to ignite advancements and foster the evolution of learning-based decision-making systems. Our ultimate goal is to contribute to a future where these systems can be safely and reliably incorporated into real-world applications, delivering greater efficiency, effectiveness, and most importantly, safety.

\clearpage
\bibliography{main}
\bibliographystyle{unsrt}

\clearpage

\appendix

\addcontentsline{toc}{section}{Appendix} 

\part{} 
\parttoc 

\section{Benchmark and Dataset Details}
\label{app:benchmark-dataset-detail}

\subsection{Hosting, Licensing, and Maintenance Plan}

Our dataset and benchmark are accessible through the following URL:
\url{www.offline-saferl.org}.
We provide three open-sourced packages, \texttt{FSRL} \footnote{\url{https://github.com/liuzuxin/FSRL}} for expert safe RL policies, \texttt{DSRL} \footnote{\url{https://github.com/liuzuxin/DSRL}} for managing datasets and environment wrappers, and \texttt{OSRL} \footnote{\url{https://github.com/liuzuxin/OSRL}} for offline safe learning algorithms.

The datasets will be hosted on our designated platform accessible via the \texttt{DSRL} package. They are also directly downloadable at \texttt{http://data.offline-saferl.org/download}. All datasets are licensed under the Creative Commons Attribution 4.0 License (CC BY). 
As for maintenance, we have established a long-term plan to keep the datasets up-to-date, correct any potential issues, and provide support to users. We also aim to further expand these datasets based on new advances in the field, thus continually promoting progress in offline safe RL research.

The \texttt{FSRL} package is under the MIT License, supporting open access and flexibility for modification and reuse. The \texttt{DSRL} package and \texttt{OSRL} package are licensed under the Apache 2.0 License, following D4RL \cite{fu2020d4rl} and CORL \cite{tarasov2022corl}. This codebase will also be regularly maintained.

\subsection{The FSRL package with safe RL experts for data collection}

The \textbf{FSRL (Fast Safe Reinforcement Learning)} package provides modularized implementations
of Safe RL algorithms based on PyTorch \cite{paszke2019pytorch} and the Tianshou framework \cite{weng2021tianshou}.
It offers high-quality and fast implementations of popular Safe RL algorithms, serving as an ideal starting point for those looking to explore and experiment in this field. By providing a comprehensive and accessible toolkit, the FSRL package aims to accelerate research in this crucial area and contribute to the development of safer and more reliable RL-powered systems. We use this package for dataset collection.

FSRL is carefully designed with several key principles in mind:

\begin{itemize}
    \item \textbf{High-quality Implementations:} Many previous safe RL implementations, such as the CPO \cite{achiam2017constrained} used by SafetyGym, have failed to meet their own benchmarks \cite{ray2019benchmarking}. This has inadvertently led to the reporting of suboptimal results in many papers that adopted these implementations. We have refined these implementations and, with suitable hyperparameters, have achieved superior safety performance in most tasks.
    \item \textbf{Fast Training Speed:} FSRL prioritizes efficient experimentation and benchmarking, offering swift training times for a variety of tasks. For instance, most algorithms can solve the SafetyCarCircle-v0 task in about 10 minutes using only 4 CPUs. 
    \item \textbf{Well-tuned Hyperparameters:} We have meticulously examined the influence of key hyperparameters for these algorithms and believe that both implementation quality and optimal hyperparameters are pivotal in creating a successful safe RL agent.
    \item \textbf{Modular Design and User-friendly Interface:} FSRL is built upon the elegant RL framework Tianshou \cite{weng2021tianshou}. It offers an agent wrapper, refactored loggers for Tensorboard and Wandb, and Pyrallis  configuration support for improved usability. Additionally, our algorithms support multiple constraints and standard RL tasks like Mujoco.
\end{itemize}

The safe RL algorithms that have been implemented in FSRL are presented in Table \ref{tab:fsrl_algos}. They include a first-order method FOCOPS \cite{zhang2020first}, a second-order method CPO \cite{achiam2017constrained}, Lagrangian-based methods \cite{stooke2020responsive}, and probabilistic-inference-based method CVPO \cite{liu2022constrained}. 

\begin{table}[ht]
\centering
\caption{\small Safe RL algorithms implemented in FSRL. \textsuperscript{1}Off-On-Policy implies that the base learning algorithm is off-policy, but the Lagrange multiplier is updated in an on-policy fashion.}
\label{tab:fsrl_algos}
\renewcommand{\arraystretch}{1.1}
\resizebox{0.95\linewidth}{!}{
\begin{tabular}{|c|c|c|}
\hline
\textbf{Algorithm} & \textbf{Type} & \textbf{Description} \\
\hline
CPO & On-Policy & Constrained Policy Optimization \cite{achiam2017constrained} \\
FOCOPS & On-Policy & First Order Constrained Optimization in Policy Space \cite{zhang2020first} \\
PPOLagrangian & On-Policy & PPO \cite{schulman2017proximal} with PID Lagrangian \cite{stooke2020responsive} \\
TRPOLagrangian & On-Policy & TRPO \cite{schulman2015trust} with PID Lagrangian \cite{stooke2020responsive} \\
DDPGLagrangian & Off-On-Policy\textsuperscript{1} & DDPG \cite{lillicrap2015continuous} with PID Lagrangian \cite{stooke2020responsive} \\
SACLagrangian & Off-On-Policy\textsuperscript{1} & SAC \cite{haarnoja2018soft} with PID Lagrangian \cite{stooke2020responsive} \\
CVPO & Off-Policy & Constrained Variational Policy Optimization \cite{liu2022constrained} \\
\hline
\end{tabular}}
\end{table}

\textbf{Dataset collection details.}
We collect datasets by training the implemented algorithms with gradually increasing or decreasing cost thresholds for each environment. By varying the algorithm, training hyper-parameters, and threshold, we are able to collect a diverse set of trajectory datasets for each task. These datasets were then merged and applied with a density filter to discard redundant trajectories with high concentrations in the cost-reward return space.

Our dataset collection process was optimized using our carefully tuned hyper-parameters and two strategies: 1) Application of a density filter to the dataset buffer during collection. This step is instrumental in preventing memory overflow. Without the filter, each task will record all collected trajectories, consuming substantial memory and leading to inefficient use of computing resources. 2) Maintenance of a relatively high minimum standard deviation ($e^{-2}$) for stochastic policies to foster exploration. This approach greatly enhances the diversity of the collected datasets. It becomes particularly relevant when cost thresholds are varied during training; an early convergence of the standard deviation may impede exploration and prevent the algorithm from adapting to new thresholds.

Details of the training configurations and hyper-parameters are available in the code.

\subsection{The DSRL package to manage the datasets and filters}

The \textbf{DSRL (Datasets for Safe Reinforcement Learning)} package serves as an offline safe RL counterpart to the widely-used D4RL. 
DSRL follows the same usage and API structure as D4RL \cite{fu2020d4rl}, making it easily accessible to researchers already familiar with D4RL. Moreover, it provides clear documentation and examples to guide users.
DSRL also ensures scalability, allowing researchers to handle large-scale datasets effectively and customize their own datasets and environments, which is crucial for testing the efficiency and scalability of their algorithms.

We provide an example code to use the dataset:


\begin{python}
import gymnasium as gym
import dsrl

env = gym.make('OfflineCarCircle-v0')

# Each task is associated with a dataset with 7 keys: 
# [observations, next_observatiosn, actions, rewards, 
# costs, terminals, timeouts]
dataset = env.get_dataset()

# An N x obs_dim Numpy array of observations
print(dataset['observations'].shape) 

# Apply dataset filters [optional]
# dataset = env.pre_process_data(dataset, filter_cfgs)

# dsrl abides by the OpenAI gym interface
obs, info = env.reset()
obs, reward, terminal, timeout, info = \ 
    env.step(env.action_space.sample())
cost = info["cost"]

\end{python}

Upon running the script, the data will be automatically fetched and stored in the \texttt{/home/user/.dsrl/datasets} directory if it does not already exist. This setup follows a similar procedure as seen in D4RL.

\subsection{Hyperparameters for OSRL algorithms}

The \textbf{OSRL (Offline Safe Reinforcement Learning)} package is a comprehensive library of offline safe RL algorithm implementations.
The framework design is inspired by the CORL \cite{tarasov2022corl} and CleanRL \cite{huang2022cleanrl} libraries, which are widely used by offline RL and online RL researchers due to their high-quality and easy-to-follow single-file implementations.
For Q-learning-based methods, we use Gaussian policies with mean vectors given as the outputs of neural networks, and with variances that are separate learnable parameters. 
The policy networks and Q networks for all experiments have two hidden layers with ReLU activation functions. 
For Lagrangian-based methods, the $K_P, K_I$ and $K_D$ are the PID parameters that control the Lagrangian multiplier. 
As for CDT, a fixed set of hyperparameters is used across all tasks. 
The majority of common parameters, including the gradient steps, remain consistent for all the methods employed. 
Each method is evaluated with three distinct cost thresholds and three random seeds. 
The primary hyperparameters employed in the experiments are summarized in Table \ref{tab:osrl-parameters}, and the comprehensive set of hyperparameters can be found in the GitHub repository.
\begin{table}[h]
\centering
\caption{Hyperparameters for \texttt{OSRL}}
\label{tab:osrl-parameters}
\renewcommand{\arraystretch}{1.2}
\resizebox{1.\linewidth}{!}{
\begin{tabular}{cccccccccc}
\cline{1-7} \cline{9-10}
Common Parameters     & \multicolumn{2}{c}{BulletGym}        & \multicolumn{2}{c}{SafetyGym}        & \multicolumn{2}{c}{MetaDrive}        &  & Parameters of CDT         & All tasks    \\ \cline{1-7} \cline{9-10} 
Actor hidden size     & \multicolumn{6}{c}{{[}256, 256{]} for all methods except CDT}                                                      &  & Number of layers          & 3            \\
VAE hidden size       & \multicolumn{6}{c}{{[}400, 400{]} BCQ-Lag, BEAR-Lag, CPQ}                                                          &  & Number of attention heads & 8            \\
Cost thresholds       & \multicolumn{2}{c}{{[}10, 20, 40{]}} & \multicolumn{2}{c}{{[}20, 40, 80{]}} & \multicolumn{2}{c}{{[}10, 20, 40{]}} &  & Embedding dimension       & 128          \\
Gradient steps        & \multicolumn{4}{c}{100000}                                                  & \multicolumn{2}{c}{200000}           &  & Batch size                & 2048         \\
{[}$K_P, K_I, K_D${]} & \multicolumn{6}{c}{{[}0.1, 0.003, 0.001{]} BCQ-Lag, BEAR-Lag}                                                      &  & Context length K          & 300          \\
Batch size            & \multicolumn{6}{c}{512}                                                                                            &  & Learning rate             & 0.0001       \\
Actor learning rate   & \multicolumn{6}{c}{0.0001}                                                                                         &  & Droupout                  & 0.1          \\
Critic learning rate  & \multicolumn{6}{c}{0.001}                                                                                          &  & Adam betas                & (0.9, 0.999) \\ \cline{1-7} \cline{9-10} 
\end{tabular}
}
\end{table}


    
    
    


\subsection{Author Responsibility Statement}

As the authors, we hereby affirm that we bear full responsibility for the datasets provided in this submission. We confirm that to the best of our knowledge, no rights are violated in the collection, distribution, and use of these datasets. They are provided under the Creative Commons Attribution 4.0 International License, which permits unrestricted use, distribution, and modification, provided appropriate credit is given to the original authors.

\section{Complete Experiment Results}

Except for the experiments for CDT, which are conducted with NVIDIA A100 GPUs, all other experiments are conducted with AMD EPYC 7542 32-Core CPUs or Intel Xeon CPUs with 4 threads. The longest experiment takes approximately one day.

\subsection{Main results for all the datasets}

We present the full experiment results on the 38 datasets in Table \ref{tab:app-complete-result}.
From the table, it can be observed that the performances of the different methods differ significantly across the tasks.
\textbf{BC-All} consistently yields high rewards but with relatively high costs often exceeding the cost threshold of 1, marking it as unsafe for most of the tasks. 
\textbf{BC-Safe} prioritizes safety and consistently delivers costs below the threshold of 1 across all tasks. While its rewards may not always be the highest, its adherence to safety makes it a reliable choice for most tasks.
\textbf{CDT} shows a more balanced performance, achieving high rewards while maintaining costs at acceptable levels for many tasks. It achieves particularly impressive results in tasks like HalfCheetahVelocity, where it attains a cost of just 0.01 while still reaping a high reward. 
Q-learning based methods (\textbf{BCQ-Lag, BEAR-Lag}, and \textbf{CPQ}) generally show a trend towards higher cost values often crossing the safety threshold, which may limit their applicability to safety-critical tasks. However, there are exceptions like in the PointGoal1 task, where BCQ-Lag shows an impressive reward while maintaining safety. 
\textbf{COptiDICE}, a stationary distribution correction-based method, demonstrates mixed results with instances of both high rewards and low costs, but also shows tasks where costs exceed the safety threshold. 

\begin{table}[ht]
\centering
\caption{\small Complete evaluation results of the normalized reward and cost. The cost threshold is 1.
The $\uparrow$ symbol denotes that the higher reward, the better. The $\downarrow$ symbol denotes that the lower cost (up to threshold 1), the better. 
Each value is averaged over 3 distinct cost thresholds, 20 evaluation episodes, and 3 random seeds.
\textbf{Bold}: Safe agents whose normalized cost is smaller than 1. 
{\color[HTML]{656565} Gray}: Unsafe agents.
{\color[HTML]{0000FF} \textbf{Blue}}: Safe agent with the highest reward.} 
\label{tab:app-complete-result}
\renewcommand{\arraystretch}{1.1}
\resizebox{1.\linewidth}{!}{
\begin{tabular}{|c|cc|cc|cc|cc|cc|cc|cc|}
\hline
                                                                  & \multicolumn{2}{c|}{BC-All}                                & \multicolumn{2}{c|}{BC-Safe}                                                 & \multicolumn{2}{c|}{CDT}                                                    & \multicolumn{2}{c|}{BCQ-Lag}                                                & \multicolumn{2}{c|}{BEAR-Lag}                              & \multicolumn{2}{c|}{CPQ}                                    & \multicolumn{2}{c|}{COptiDICE}                                              \\ \cline{2-15} 
\multirow{-2}{*}{Task}                                            & reward $\uparrow$                       & cost $\downarrow$                        & reward $\uparrow$                                & cost $\downarrow$                                 & reward $\uparrow$                               & cost $\downarrow$                                 & reward $\uparrow$                               & cost $\downarrow$                                 & reward $\uparrow$                       & cost $\downarrow$                        & reward $\uparrow$                       & cost $\downarrow$                         & reward $\uparrow$                               & cost $\downarrow$                                 \\ \hline
PointButton1                                                         & {\color[HTML]{656565} 0.1}   & {\color[HTML]{656565} 1.05} & {\color[HTML]{0000FF} \textbf{0.06}}  & {\color[HTML]{0000FF} \textbf{0.52}} & {\color[HTML]{656565} 0.53}          & {\color[HTML]{656565} 1.68}          & {\color[HTML]{656565} 0.24}          & {\color[HTML]{656565} 1.73}          & {\color[HTML]{656565} 0.2}   & {\color[HTML]{656565} 1.6}  & {\color[HTML]{656565} 0.69}          & {\color[HTML]{656565} 3.2}           & {\color[HTML]{656565} 0.13}          & {\color[HTML]{656565} 1.35}          \\
PointButton2                                                         & {\color[HTML]{656565} 0.27}  & {\color[HTML]{656565} 2.02} & {\color[HTML]{656565} 0.16}           & {\color[HTML]{656565} 1.1}           & {\color[HTML]{656565} 0.46}          & {\color[HTML]{656565} 1.57}          & {\color[HTML]{656565} 0.4}           & {\color[HTML]{656565} 2.66}          & {\color[HTML]{656565} 0.43}  & {\color[HTML]{656565} 2.47} & {\color[HTML]{656565} 0.58}          & {\color[HTML]{656565} 4.3}           & {\color[HTML]{656565} 0.15}          & {\color[HTML]{656565} 1.51}          \\
PointCircle1                                                         & {\color[HTML]{656565} 0.79}  & {\color[HTML]{656565} 3.98} & \textbf{0.41}                         & \textbf{0.16}                        & {\color[HTML]{0000FF} \textbf{0.59}} & {\color[HTML]{0000FF} \textbf{0.69}} & {\color[HTML]{656565} 0.54}          & {\color[HTML]{656565} 2.38}          & {\color[HTML]{656565} 0.73}  & {\color[HTML]{656565} 3.28} & \textbf{0.43}                        & \textbf{0.75}                        & {\color[HTML]{656565} 0.86}          & {\color[HTML]{656565} 5.51}          \\
PointCircle2                                                         & {\color[HTML]{656565} 0.66}  & {\color[HTML]{656565} 4.17} & {\color[HTML]{0000FF} \textbf{0.48}}  & {\color[HTML]{0000FF} \textbf{0.99}} & {\color[HTML]{656565} 0.64}          & {\color[HTML]{656565} 1.05}          & {\color[HTML]{656565} 0.66}          & {\color[HTML]{656565} 2.6}           & {\color[HTML]{656565} 0.63}  & {\color[HTML]{656565} 4.27} & {\color[HTML]{656565} 0.24}          & {\color[HTML]{656565} 3.58}          & {\color[HTML]{656565} 0.85}          & {\color[HTML]{656565} 8.61}          \\
PointGoal1                                                           & \textbf{0.65}                & \textbf{0.95}               & \textbf{0.43}                         & \textbf{0.54}                        & {\color[HTML]{656565} 0.69}          & {\color[HTML]{656565} 1.12}          & {\color[HTML]{0000FF} \textbf{0.71}} & {\color[HTML]{0000FF} \textbf{0.98}} & 0.74                         & 1.18                        & \textbf{0.57}                        & \textbf{0.35}                        & {\color[HTML]{656565} 0.49}          & {\color[HTML]{656565} 1.66}          \\
PointGoal2                                                           & {\color[HTML]{656565} 0.54}  & {\color[HTML]{656565} 1.97} & {\color[HTML]{0000FF} \textbf{0.29}}  & {\color[HTML]{0000FF} \textbf{0.78}} & {\color[HTML]{656565} 0.59}          & {\color[HTML]{656565} 1.34}          & {\color[HTML]{656565} 0.67}          & {\color[HTML]{656565} 3.18}          & {\color[HTML]{656565} 0.67}  & {\color[HTML]{656565} 3.11} & {\color[HTML]{656565} 0.4}           & {\color[HTML]{656565} 1.31}          & {\color[HTML]{656565} 0.38}          & {\color[HTML]{656565} 1.92}          \\
PointPush1                                                           & \textbf{0.19}                & \textbf{0.61}               & \textbf{0.13}                         & \textbf{0.43}                        & \textbf{0.24}                        & \textbf{0.48}                        & {\color[HTML]{0000FF} \textbf{0.33}} & {\color[HTML]{0000FF} \textbf{0.86}} & \textbf{0.22}                & \textbf{0.79}               & \textbf{0.2}                         & \textbf{0.83}                        & \textbf{0.13}                        & \textbf{0.83}                        \\
PointPush2                                                           & \textbf{0.18}                & \textbf{0.91}               & \textbf{0.11}                         & \textbf{0.8}                         & \textbf{0.21}                        & \textbf{0.65}                        & {\color[HTML]{0000FF} \textbf{0.23}} & {\color[HTML]{0000FF} \textbf{0.99}} & \textbf{0.16}                & \textbf{0.89}               & 0.11                                 & 1.04                                 & 0.02                                 & 1.18                                 \\
CarButton1                                                           & {\color[HTML]{656565} 0.03}  & {\color[HTML]{656565} 1.38} & {\color[HTML]{0000FF} \textbf{0.07}}  & {\color[HTML]{0000FF} \textbf{0.85}} & {\color[HTML]{656565} 0.21}          & {\color[HTML]{656565} 1.6}           & {\color[HTML]{656565} 0.04}          & {\color[HTML]{656565} 1.63}          & {\color[HTML]{656565} 0.18}  & {\color[HTML]{656565} 2.72} & {\color[HTML]{656565} 0.42}          & {\color[HTML]{656565} 9.66}          & {\color[HTML]{656565} -0.08}         & {\color[HTML]{656565} 1.68}          \\
CarButton2                                                           & {\color[HTML]{656565} -0.13} & {\color[HTML]{656565} 1.24} & {\color[HTML]{0000FF} \textbf{-0.01}} & {\color[HTML]{0000FF} \textbf{0.63}} & {\color[HTML]{656565} 0.13}          & {\color[HTML]{656565} 1.58}          & {\color[HTML]{656565} 0.06}          & {\color[HTML]{656565} 2.13}          & {\color[HTML]{656565} -0.01} & {\color[HTML]{656565} 2.29} & {\color[HTML]{656565} 0.37}          & {\color[HTML]{656565} 12.51}         & {\color[HTML]{656565} -0.07}         & {\color[HTML]{656565} 1.59}          \\
CarCircle1                                                           & {\color[HTML]{656565} 0.72}  & {\color[HTML]{656565} 4.39} & {\color[HTML]{656565} 0.37}           & {\color[HTML]{656565} 1.38}          & {\color[HTML]{656565} 0.6}           & {\color[HTML]{656565} 1.73}          & {\color[HTML]{656565} 0.73}          & {\color[HTML]{656565} 5.25}          & {\color[HTML]{656565} 0.76}  & {\color[HTML]{656565} 5.46} & {\color[HTML]{656565} 0.02}          & {\color[HTML]{656565} 2.29}          & {\color[HTML]{656565} 0.7}           & {\color[HTML]{656565} 5.72}          \\
CarCircle2                                                           & {\color[HTML]{656565} 0.76}  & {\color[HTML]{656565} 6.44} & {\color[HTML]{656565} 0.54}           & {\color[HTML]{656565} 3.38}          & {\color[HTML]{656565} 0.66}          & {\color[HTML]{656565} 2.53}          & {\color[HTML]{656565} 0.72}          & {\color[HTML]{656565} 6.58}          & {\color[HTML]{656565} 0.74}  & {\color[HTML]{656565} 6.82} & {\color[HTML]{656565} 0.44}          & {\color[HTML]{656565} 2.69}          & {\color[HTML]{656565} 0.77}          & {\color[HTML]{656565} 7.99}          \\
CarGoal1                                                             & \textbf{0.39}                & \textbf{0.33}               & \textbf{0.24}                         & \textbf{0.28}                        & {\color[HTML]{656565} 0.66}          & {\color[HTML]{656565} 1.21}          & {\color[HTML]{0000FF} \textbf{0.47}} & {\color[HTML]{0000FF} \textbf{0.78}} & {\color[HTML]{656565} 0.61}  & {\color[HTML]{656565} 1.13} & {\color[HTML]{656565} 0.79}          & {\color[HTML]{656565} 1.42}          & \textbf{0.35}                        & \textbf{0.54}                        \\
CarGoal2                                                             & {\color[HTML]{656565} 0.23}  & {\color[HTML]{656565} 1.05} & \textbf{0.14}                         & \textbf{0.51}                        & {\color[HTML]{656565} 0.48}          & {\color[HTML]{656565} 1.25}          & {\color[HTML]{656565} 0.3}           & {\color[HTML]{656565} 1.44}          & {\color[HTML]{656565} 0.28}  & {\color[HTML]{656565} 1.01} & {\color[HTML]{656565} 0.65}          & {\color[HTML]{656565} 3.75}          & {\color[HTML]{0000FF} \textbf{0.25}} & {\color[HTML]{0000FF} \textbf{0.91}} \\
CarPush1                                                             & \textbf{0.22}                & \textbf{0.36}               & \textbf{0.14}                         & \textbf{0.33}                        & {\color[HTML]{0000FF} \textbf{0.31}} & {\color[HTML]{0000FF} \textbf{0.4}}  & \textbf{0.23}                        & \textbf{0.43}                        & \textbf{0.21}                & \textbf{0.54}               & \textbf{-0.03}                       & \textbf{0.95}                        & \textbf{0.23}                        & \textbf{0.5}                         \\
CarPush2                                                             & \textbf{0.14}                & \textbf{0.9}                & \textbf{0.05}                         & \textbf{0.45}                        & {\color[HTML]{656565} 0.19}          & {\color[HTML]{656565} 1.3}           & {\color[HTML]{656565} 0.15}          & {\color[HTML]{656565} 1.38}          & {\color[HTML]{656565} 0.1}   & {\color[HTML]{656565} 1.2}  & {\color[HTML]{656565} 0.24}          & {\color[HTML]{656565} 4.25}          & {\color[HTML]{656565} 0.09}          & {\color[HTML]{656565} 1.07}          \\
SwimmerVelocity                                                      & {\color[HTML]{656565} 0.49}  & {\color[HTML]{656565} 4.72} & {\color[HTML]{656565} 0.51}           & {\color[HTML]{656565} 1.07}          & {\color[HTML]{0000FF} \textbf{0.66}} & {\color[HTML]{0000FF} \textbf{0.96}} & {\color[HTML]{656565} 0.48}          & {\color[HTML]{656565} 6.58}          & {\color[HTML]{656565} 0.3}   & {\color[HTML]{656565} 2.33} & {\color[HTML]{656565} 0.13}          & {\color[HTML]{656565} 2.66}          & {\color[HTML]{656565} 0.63}          & {\color[HTML]{656565} 7.58}          \\
HopperVelocity                                                       & {\color[HTML]{656565} 0.65}  & {\color[HTML]{656565} 6.39} & \textbf{0.36}                         & \textbf{0.67}                        & \textbf{0.63}                        & \textbf{0.61}                        & {\color[HTML]{656565} 0.78}          & {\color[HTML]{656565} 5.02}          & {\color[HTML]{656565} 0.34}  & {\color[HTML]{656565} 5.86} & {\color[HTML]{656565} 0.14}          & {\color[HTML]{656565} 2.11}          & {\color[HTML]{656565} 0.13}          & {\color[HTML]{656565} 1.51}          \\
HalfCheetahVelocity                                                  & {\color[HTML]{656565} 0.97}  & {\color[HTML]{656565} 13.1} & \textbf{0.88}                         & \textbf{0.54}                        & {\color[HTML]{0000FF} \textbf{1.0}}  & {\color[HTML]{0000FF} \textbf{0.01}} & {\color[HTML]{656565} 1.05}          & {\color[HTML]{656565} 18.21}         & {\color[HTML]{656565} 0.98}  & {\color[HTML]{656565} 6.58} & \textbf{0.29}                        & \textbf{0.74}                        & \textbf{0.65}                        & \textbf{0.0}                         \\
Walker2dVelocity                                                     & {\color[HTML]{656565} 0.79}  & {\color[HTML]{656565} 3.88} & {\color[HTML]{0000FF} \textbf{0.79}}  & {\color[HTML]{0000FF} \textbf{0.04}} & \textbf{0.78}                        & \textbf{0.06}                        & \textbf{0.79}                        & \textbf{0.17}                        & {\color[HTML]{656565} 0.86}  & {\color[HTML]{656565} 3.1}  & {\color[HTML]{656565} \textbf{0.04}} & {\color[HTML]{656565} \textbf{0.21}} & {\color[HTML]{656565} \textbf{0.12}} & {\color[HTML]{656565} \textbf{0.74}} \\
AntVelocity                                                          & {\color[HTML]{656565} 0.98}  & {\color[HTML]{656565} 3.72} & {\color[HTML]{0000FF} \textbf{0.98}}  & {\color[HTML]{0000FF} \textbf{0.29}} & \textbf{0.98}                        & \textbf{0.39}                        & {\color[HTML]{656565} 1.02}          & {\color[HTML]{656565} 4.15}          & \textbf{-1.01}               & \textbf{0.0}                & \textbf{-1.01}                       & \textbf{0.0}                         & {\color[HTML]{656565} 1.0}           & {\color[HTML]{656565} 3.28}          \\ \hline
\textbf{\begin{tabular}[c]{@{}c@{}}SafetyGym\\ Average\end{tabular}} & {\color[HTML]{656565} 0.46}  & {\color[HTML]{656565} 3.03} & {\color[HTML]{0000FF} \textbf{0.34}}  & {\color[HTML]{0000FF} \textbf{0.75}} & {\color[HTML]{656565} 0.54}          & {\color[HTML]{656565} 1.06}          & {\color[HTML]{656565} 0.5}           & {\color[HTML]{656565} 3.29}          & {\color[HTML]{656565} 0.39}  & {\color[HTML]{656565} 2.7}  & {\color[HTML]{656565} 0.27}          & {\color[HTML]{656565} 2.79}          & {\color[HTML]{656565} 0.37}          & {\color[HTML]{656565} 2.65}          \\ \hline
BallRun                                                              & {\color[HTML]{656565} 0.6}   & {\color[HTML]{656565} 5.08} & {\color[HTML]{656565} 0.27}           & {\color[HTML]{656565} 1.46}          & {\color[HTML]{656565} 0.39}          & {\color[HTML]{656565} 1.16}          & {\color[HTML]{656565} 0.76}          & {\color[HTML]{656565} 3.91}          & {\color[HTML]{656565} -0.47} & {\color[HTML]{656565} 5.03} & {\color[HTML]{656565} 0.22}          & {\color[HTML]{656565} 1.27}          & {\color[HTML]{656565} 0.59}          & {\color[HTML]{656565} 3.52}          \\
CarRun                                                               & \textbf{0.97}                & \textbf{0.33}               & \textbf{0.94}                         & \textbf{0.22}                        & {\color[HTML]{0000FF} \textbf{0.99}} & {\color[HTML]{0000FF} \textbf{0.65}} & \textbf{0.94}                        & \textbf{0.15}                        & {\color[HTML]{656565} 0.68}  & {\color[HTML]{656565} 7.78} & {\color[HTML]{656565} 0.95}          & {\color[HTML]{656565} 1.79}          & \textbf{0.87}                        & \textbf{0.0}                         \\
DroneRun                                                             & {\color[HTML]{656565} 0.24}  & {\color[HTML]{656565} 2.13} & \textbf{0.28}                         & \textbf{0.74}                        & {\color[HTML]{0000FF} \textbf{0.63}} & {\color[HTML]{0000FF} \textbf{0.79}} & {\color[HTML]{656565} 0.72}          & {\color[HTML]{656565} 5.54}          & {\color[HTML]{656565} 0.42}  & {\color[HTML]{656565} 2.47} & {\color[HTML]{656565} 0.33}          & {\color[HTML]{656565} 3.52}          & {\color[HTML]{656565} 0.67}          & {\color[HTML]{656565} 4.15}          \\
AntRun                                                               & {\color[HTML]{656565} 0.72}  & {\color[HTML]{656565} 2.93} & {\color[HTML]{656565} 0.65}           & {\color[HTML]{656565} 1.09}          & {\color[HTML]{0000FF} \textbf{0.72}} & {\color[HTML]{0000FF} \textbf{0.91}} & {\color[HTML]{656565} 0.76}          & {\color[HTML]{656565} 5.11}          & \textbf{0.15}                & \textbf{0.73}               & \textbf{0.03}                        & \textbf{0.02}                        & \textbf{0.61}                        & \textbf{0.94}                        \\
BallCircle                                                           & {\color[HTML]{656565} 0.74}  & {\color[HTML]{656565} 4.71} & \textbf{0.52}                         & \textbf{0.65}                        & {\color[HTML]{656565} 0.77}          & {\color[HTML]{656565} 1.07}          & {\color[HTML]{656565} 0.69}          & {\color[HTML]{656565} 2.36}          & {\color[HTML]{656565} 0.86}  & {\color[HTML]{656565} 3.09} & {\color[HTML]{0000FF} \textbf{0.64}} & {\color[HTML]{0000FF} \textbf{0.76}} & {\color[HTML]{656565} 0.7}           & {\color[HTML]{656565} 2.61}          \\
CarCircle                                                            & {\color[HTML]{656565} 0.58}  & {\color[HTML]{656565} 3.74} & \textbf{0.5}                          & \textbf{0.84}                        & {\color[HTML]{0000FF} \textbf{0.75}} & {\color[HTML]{0000FF} \textbf{0.95}} & {\color[HTML]{656565} 0.63}          & {\color[HTML]{656565} 1.89}          & {\color[HTML]{656565} 0.74}  & {\color[HTML]{656565} 2.18} & \textbf{0.71}                        & \textbf{0.33}                        & {\color[HTML]{656565} 0.49}          & {\color[HTML]{656565} 3.14}          \\
DroneCircle                                                          & {\color[HTML]{656565} 0.72}  & {\color[HTML]{656565} 3.03} & \textbf{0.56}                         & \textbf{0.57}                        & {\color[HTML]{0000FF} \textbf{0.63}} & {\color[HTML]{0000FF} \textbf{0.98}} & {\color[HTML]{656565} 0.8}           & {\color[HTML]{656565} 3.07}          & {\color[HTML]{656565} 0.78}  & {\color[HTML]{656565} 3.68} & {\color[HTML]{656565} -0.22}         & {\color[HTML]{656565} 1.28}          & {\color[HTML]{656565} 0.26}          & {\color[HTML]{656565} 1.02}          \\
AntCircle                                                            & {\color[HTML]{656565} 0.58}  & {\color[HTML]{656565} 4.9}  & \textbf{0.4}                          & \textbf{0.96}                        & {\color[HTML]{0000FF} \textbf{0.54}} & {\color[HTML]{0000FF} \textbf{1.78}} & {\color[HTML]{656565} 0.58}          & {\color[HTML]{656565} 2.87}          & {\color[HTML]{656565} 0.65}  & {\color[HTML]{656565} 5.48} & \textbf{0.0}                         & \textbf{0.0}                         & {\color[HTML]{656565} 0.17}          & {\color[HTML]{656565} 5.04}          \\ \hline
\textbf{\begin{tabular}[c]{@{}c@{}}BulletGym\\ Average\end{tabular}} & {\color[HTML]{656565} 0.64}  & {\color[HTML]{656565} 3.36} & {\color[HTML]{0000FF} \textbf{0.52}}  & {\color[HTML]{0000FF} \textbf{0.82}} & {\color[HTML]{656565} 0.68}          & {\color[HTML]{656565} 1.04}          & {\color[HTML]{656565} 0.74}          & {\color[HTML]{656565} 3.11}          & {\color[HTML]{656565} 0.48}  & {\color[HTML]{656565} 3.8}  & {\color[HTML]{656565} 0.33}          & {\color[HTML]{656565} 1.12}          & {\color[HTML]{656565} 0.55}          & {\color[HTML]{656565} 2.55}          \\ \hline
easysparse                                                           & {\color[HTML]{656565} 0.17}  & {\color[HTML]{656565} 1.54} & \textbf{0.11}                         & \textbf{0.21}                        & {\color[HTML]{0000FF} \textbf{0.17}} & {\color[HTML]{0000FF} \textbf{0.23}} & {\color[HTML]{656565} 0.78}          & {\color[HTML]{656565} 5.01}          & \textbf{0.11}                & \textbf{0.86}               & \textbf{-0.06}                       & \textbf{0.07}                        & {\color[HTML]{656565} 0.96}          & {\color[HTML]{656565} 5.44}          \\
eastmean                                                             & {\color[HTML]{656565} 0.43}  & {\color[HTML]{656565} 2.82} & \textbf{0.04}                         & \textbf{0.29}                        & {\color[HTML]{0000FF} \textbf{0.45}} & {\color[HTML]{0000FF} \textbf{0.54}} & {\color[HTML]{656565} 0.71}          & {\color[HTML]{656565} 3.44}          & \textbf{0.08}                & \textbf{0.86}               & \textbf{-0.07}                       & \textbf{0.07}                        & {\color[HTML]{656565} 0.66}          & {\color[HTML]{656565} 3.97}          \\
easydense                                                            & {\color[HTML]{656565} 0.27}  & {\color[HTML]{656565} 1.94} & \textbf{0.11}                         & \textbf{0.14}                        & {\color[HTML]{0000FF} \textbf{0.32}} & {\color[HTML]{0000FF} \textbf{0.62}} & \textbf{0.26}                        & \textbf{0.47}                        & \textbf{0.02}                & \textbf{0.41}               & \textbf{-0.06}                       & \textbf{0.03}                        & {\color[HTML]{656565} 0.5}           & {\color[HTML]{656565} 2.54}          \\
mediumsparse                                                         & {\color[HTML]{656565} 0.83}  & {\color[HTML]{656565} 3.34} & {\color[HTML]{0000FF} \textbf{0.33}}  & {\color[HTML]{0000FF} \textbf{0.3}}  & {\color[HTML]{656565} 0.87}          & {\color[HTML]{656565} 1.1}           & {\color[HTML]{656565} 0.44}          & {\color[HTML]{656565} 1.16}          & \textbf{-0.03}               & \textbf{0.17}               & \textbf{-0.08}                       & \textbf{0.07}                        & {\color[HTML]{656565} 0.71}          & {\color[HTML]{656565} 2.49}          \\
mediummean                                                           & {\color[HTML]{656565} 0.77}  & {\color[HTML]{656565} 2.53} & \textbf{0.31}                         & \textbf{0.21}                        & {\color[HTML]{0000FF} \textbf{0.45}} & {\color[HTML]{0000FF} \textbf{0.75}} & {\color[HTML]{656565} 0.78}          & {\color[HTML]{656565} 1.53}          & \textbf{-0.0}                & \textbf{0.34}               & \textbf{-0.08}                       & \textbf{0.05}                        & {\color[HTML]{656565} 0.76}          & {\color[HTML]{656565} 2.05}          \\
mediumdense                                                          & {\color[HTML]{656565} 0.45}  & {\color[HTML]{656565} 1.47} & {\color[HTML]{0000FF} \textbf{0.24}}  & {\color[HTML]{0000FF} \textbf{0.17}} & {\color[HTML]{656565} 0.88}          & {\color[HTML]{656565} 2.41}          & {\color[HTML]{656565} 0.58}          & {\color[HTML]{656565} 1.89}          & \textbf{0.01}                & \textbf{0.28}               & \textbf{-0.07}                       & \textbf{0.07}                        & {\color[HTML]{656565} 0.69}          & {\color[HTML]{656565} 2.24}          \\
hardsparse                                                           & {\color[HTML]{656565} 0.42}  & {\color[HTML]{656565} 1.8}  & {\color[HTML]{656565} 0.17}           & {\color[HTML]{656565} 3.25}          & {\color[HTML]{0000FF} \textbf{0.25}} & {\color[HTML]{0000FF} \textbf{0.41}} & {\color[HTML]{656565} 0.5}           & {\color[HTML]{656565} 1.02}          & \textbf{0.01}                & \textbf{0.16}               & \textbf{-0.05}                       & \textbf{0.06}                        & {\color[HTML]{656565} 0.37}          & {\color[HTML]{656565} 2.05}          \\
hardmean                                                             & {\color[HTML]{656565} 0.2}   & {\color[HTML]{656565} 1.77} & \textbf{0.13}                         & \textbf{0.4}                         & {\color[HTML]{0000FF} \textbf{0.33}} & {\color[HTML]{0000FF} \textbf{0.97}} & {\color[HTML]{656565} 0.47}          & {\color[HTML]{656565} 2.56}          & \textbf{-0.0}                & \textbf{0.21}               & \textbf{-0.05}                       & \textbf{0.06}                        & {\color[HTML]{656565} 0.32}          & {\color[HTML]{656565} 2.47}          \\
harddense                                                            & {\color[HTML]{656565} 0.2}   & {\color[HTML]{656565} 1.33} & {\color[HTML]{0000FF} \textbf{0.15}}  & {\color[HTML]{0000FF} \textbf{0.22}} & \textbf{0.08}                        & \textbf{0.21}                        & {\color[HTML]{656565} 0.35}          & {\color[HTML]{656565} 1.4}           & \textbf{0.02}                & \textbf{0.26}               & \textbf{-0.04}                       & \textbf{0.08}                        & {\color[HTML]{656565} 0.24}          & {\color[HTML]{656565} 1.68}          \\ \hline
\textbf{\begin{tabular}[c]{@{}c@{}}MetaDrive\\ Average\end{tabular}} & {\color[HTML]{656565} 0.42}  & {\color[HTML]{656565} 2.06} & \textbf{0.18}                         & \textbf{0.58}                        & {\color[HTML]{0000FF} \textbf{0.42}} & {\color[HTML]{0000FF} \textbf{0.8}}  & {\color[HTML]{656565} 0.54}          & {\color[HTML]{656565} 2.05}          & \textbf{0.02}                & \textbf{0.39}               & \textbf{-0.06}                       & \textbf{0.06}                        & {\color[HTML]{656565} 0.58}          & {\color[HTML]{656565} 2.77}          \\ \hline

\end{tabular}
}
\end{table}

The disparity observed in the performance of learning algorithms across different simulation environments underscores the influence of task definition on algorithmic efficacy. For instance, while numerous methods struggle to maintain safety in Safety Gymnasium's Circle tasks, the same algorithms can maintain safety in BulletSafetyGym's Circle tasks. Interestingly, their task definitions are almost identical. This divergence in outcomes can be attributed to the different time horizons and simulation steps, where SafetyGymnasium tasks typically feature longer time horizons with shorter simulation steps per iteration, while BulletSafetyGym tasks have shorter time horizons, which could potentially facilitate training. Hence, task design crucially impacts algorithmic performance, suggesting that future research should focus on the relationship between task definition, including CMDP design, and algorithm efficacy to foster safer decision-making systems.

Overall, the choice of the best method highly depends on the specific task and whether safety or reward maximization is the primary goal. While BC-Safe tends to be the safest, CDT and COptiDICE can sometimes achieve a good balance between safety and reward. In some tasks, Q-learning based methods might also be the best choice when costs are within acceptable boundaries.

\subsection{Post-process filtering results}

We present the complete results for the noisy level manipulation filter and the partial data discarding filter in Sec. \ref{sec:post-process} as follows.
Each value is averaged among the corresponding Bullet-Gym and Safety-Gym suites of tasks for better illustration. 
We detail each filter as follows.

\textbf{Noise-level manipulation filter.} The real-world dataset could be noisy and contains trajectories that accidentally record high reward returns and small cost returns despite following a poor behavioral policy. 
To evaluate the training robustness of offline learning algorithms, we use this filter to create such datasets that contain different portions ($\alpha \% = 0\%, 1\%,5\%,10\%$) of outlier trajectories. 
We consider the task with stochastic reward and cost function, i.e., high-cost trajectories have the probability of $\alpha\%$ to be labeled as a ``lucky" trajectory with high reward and low cost. 
Specifically, we select $\alpha \%$ high-cost trajectories and modify their cost return to be less than the cost threshold and their reward returns to be high.

\begin{figure}[h]
\vspace{-1mm}
    \centering    \includegraphics[width=0.99\linewidth]{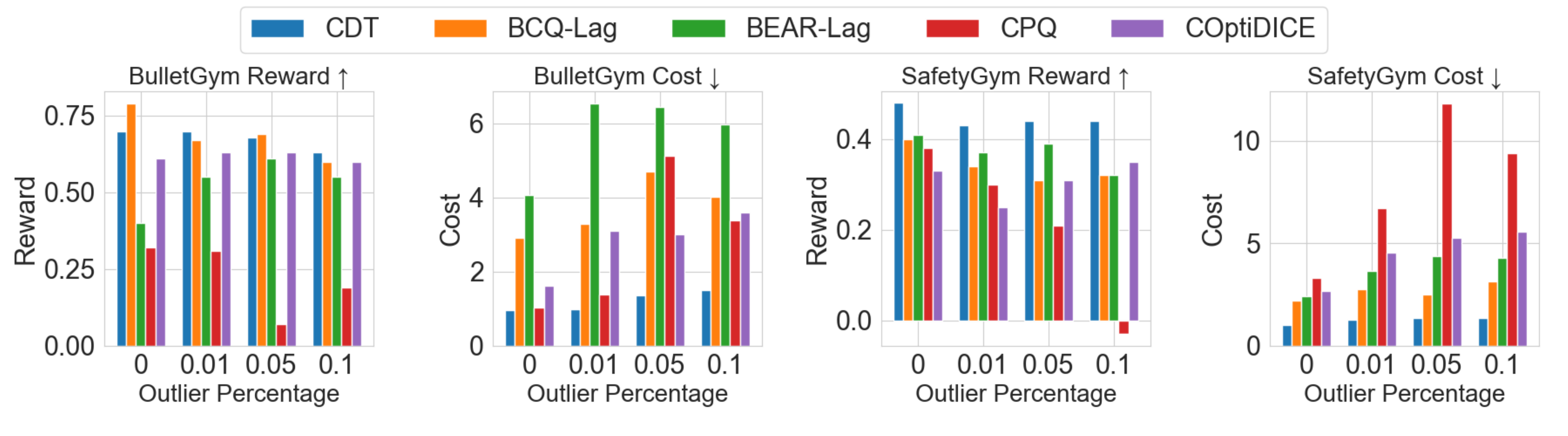}
    \caption{\small Average performance with different percentage of outlier trajectories.}
    \label{fig:noise-app}
\end{figure}

Fig. \ref{fig:noise-app} illustrates the performance impact of the noise data filter, which increases the percentage of outlier trajectories. 
This plot is a corrected version of the one in the main content.
Several algorithms, such as CPQ in SafetyGym tasks, exhibit drastic performance degradation with a substantial reward reduction and cost increase. This phenomenon underscores the critical importance of outlier sensitivity in the evaluation of offline safe RL algorithms. Excessive sensitivity to outliers may introduce instability during the learning process, leading to suboptimal safety and performance outcomes.

\textbf{Partial data discarding filter.}
The datasets could hardly be perfect and contain all different situations, i.e., with complete coverage of all possible reward and cost returns. The real-world data could be either overly conservative or overly risky. This filter mimics this by selectively discarding trajectories within defined return ranges, which helps us gauge an algorithm's ability to handle unseen safety thresholds and learn from sub-optimal data. 

\begin{figure}[h]
\vspace{-1mm}
    \centering    \includegraphics[width=0.99\linewidth]{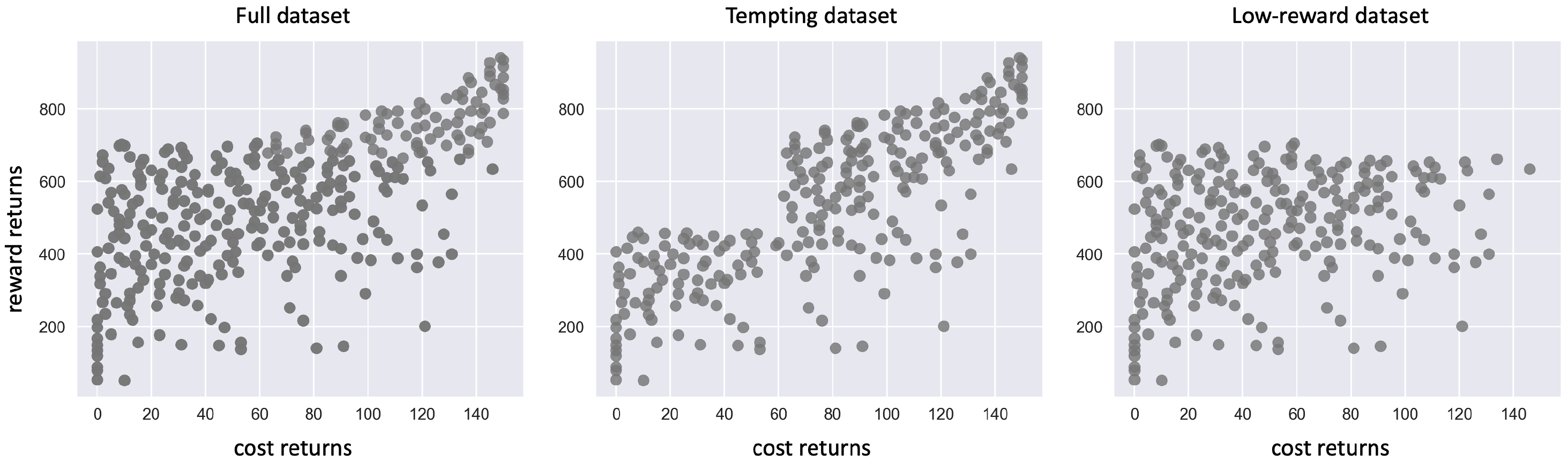}
    \vspace{-1mm}
    \caption{\small Illustration of the partial data discarding filters.}
    \label{fig:partial_data_filter}
    \vspace{-1mm}
\end{figure}

More specifically, we adopt two data discarding strategies: 1) discarding the top $50\%-100\%$ reward return trajectories within the $0\%-50\%$ cost return regions to create a \textbf{tempting} dataset \cite{liu2022robustness}; 2) discarding the top $50\%-100\%$ reward return trajectories within the $50\%-100\%$ cost return regions to create a \textbf{low-reward} and sub-optimal dataset.
The first dataset is tempting because it contains high-reward and high-cost trajectories that could potentially lead the agent to pursue risky behaviors. The learning algorithm must balance high-reward-high-cost and low-reward-low-cost performance, posing more challenges than the full datasets.
On the contrary, the second \textbf{low-reward} dataset could lead to a conservative learned policy. Though the trained agent is safe, the reward could also be low. The tempting dataset and low-reward dataset are shown in Figure~\ref{fig:partial_data_filter}.

\begin{figure}[h]
    \centering    \includegraphics[width=0.99\linewidth]{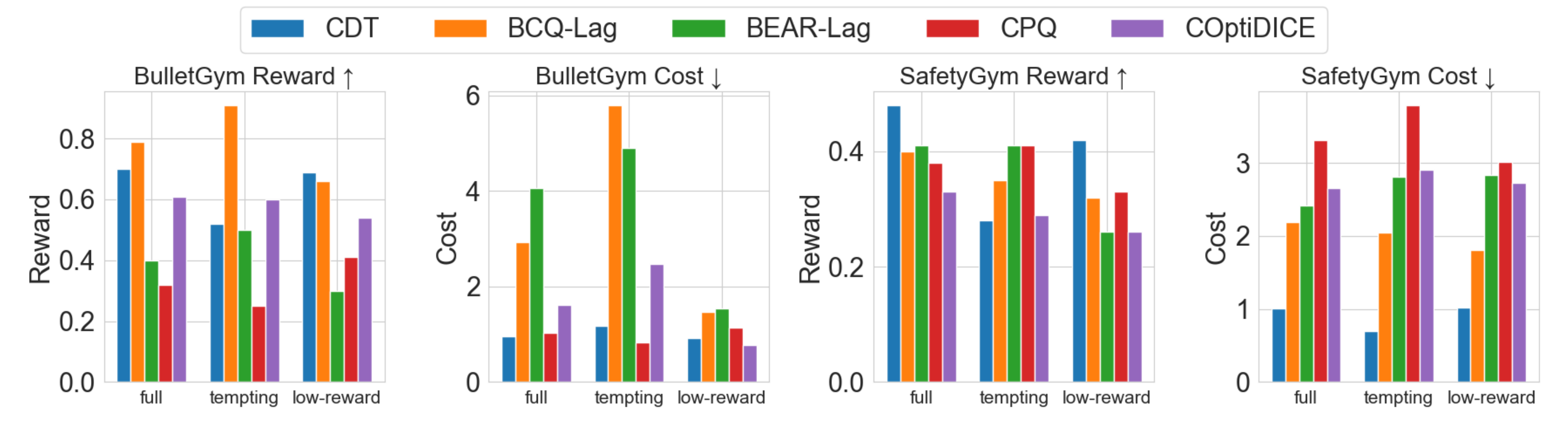}
    \vspace{-1mm}
    \caption{\small Average performance with different data discarding strategy.}
    \label{fig:data-discard}
    \vspace{-1mm}
\end{figure}

Figure \ref{fig:data-discard} shows the performance under different data-discarding strategies. We can see that The tempting datasets usually lead to high costs and high rewards, i.e., tempting policies, in the BulletSafetyGym tasks. While the low-reward datasets tend to reduce the cost, making the learning algorithm safer. These results consolidate the inherent trade-offs between the reward and cost in the safe learning problem. It also provides an insight that we can adjust the learning difficulty by manipulating the shape of the datasets. Investigating how to selectively use data in the dataset for learning to enhance safety and performance could be an interesting future direction.

\section{Dataset Documentation and Visualization}

\subsection{Dataset Breakdown Details and Intended Uses}

The datasets included in this submission are intended for use in the research and development of offline safe learning algorithms. They are diverse, encompassing three different safe RL environments and are designed to test algorithms on a range of safety thresholds. 
Documentation of the dataset, including a detailed breakdown of environments, tasks, and data sizes, can be found in the following Table \ref{tab:dataset-details}.
The \texttt{Max Cost} column denotes the maximum cost return in the dataset trajectories.

\begin{table}[ht]
\centering
\caption{\small Dataset details. The Max Cost column means the maximum cost return in the dataset trajectories.} 
\label{tab:dataset-details}
\renewcommand{\arraystretch}{1.1}
\resizebox{0.99\linewidth}{!}{
\begin{tabular}{|c|c|c|c|c|c|c|}
\hline
Benchmarks                        & Task                          & Max Timestep & Action Space & State Space & Max Cost & Trajectories \\ \hline
\multirow{21}{*}{SafetyGymnasium} & SafetyPointGoal1-v0           & 1000         & 2            & 60          & 100      & 2022         \\
                                  & SafetyPointGoal2-v0           & 1000         & 2            & 60          & 200      & 3442         \\
                                  & SafetyPointButton1-v0         & 1000         & 2            & 76          & 200      & 2268         \\
                                  & SafetyPointButton2-v0         & 1000         & 2            & 76          & 250      & 3288         \\
                                  & SafetyPointPush1-v0           & 1000         & 2            & 76          & 150      & 2379         \\
                                  & SafetyPointPush2-v0           & 1000         & 2            & 76          & 200      & 3242         \\
                                  & SafetyPointCircle1-v0         & 500          & 2            & 28          & 200      & 1098         \\
                                  & SafetyPointCircle2-v0         & 500          & 2            & 28          & 300      & 895          \\
                                  & SafetyCarGoal1-v0             & 1000         & 2            & 72          & 120      & 1671         \\
                                  & SafetyCarGoal2-v0             & 1000         & 2            & 72          & 200      & 4105         \\
                                  & SafetyCarButton1-v0           & 1000         & 2            & 88          & 250      & 2656         \\
                                  & SafetyCarButton2-v0           & 1000         & 2            & 88          & 300      & 3755         \\
                                  & SafetyCarPush1-v0             & 1000         & 2            & 88          & 200      & 2871         \\
                                  & SafetyCarPush2-v0             & 1000         & 2            & 88          & 250      & 4407         \\
                                  & SafetyCarCircle1-v0           & 500          & 2            & 40          & 250      & 1271         \\
                                  & SafetyCarCircle2-v0           & 500          & 2            & 40          & 400      & 940          \\
                                  & SafetySwimmerVelocity-v1      & 1000         & 2            & 8           & 200      & 1686         \\
                                  & SafetyHopperVelocity-v1       & 1000         & 3            & 11          & 250      & 2240         \\
                                  & SafetyHalfCheetahVelocity-v1  & 1000         & 6            & 17          & 250      & 2495         \\
                                  & SafetyWalker2dVelocity-v1     & 1000         & 6            & 17          & 300      & 2729         \\
                                  & SafetyAntVelocity-v1          & 1000         & 8            & 27          & 250      & 2249         \\ \hline
\multirow{8}{*}{BulletSafetyGym}  & SafetyBallRun-v0              & 100          & 2            & 7           & 80       & 940          \\
                                  & SafetyCarRun-v0               & 200          & 2            & 7           & 40       & 651          \\
                                  & SafetyDroneRun-v0             & 200          & 4            & 17          & 140      & 1990         \\
                                  & SafetyAntRun-v0               & 200          & 8            & 33          & 150      & 1816         \\
                                  & SafetyBallCircle-v0           & 200          & 2            & 8           & 80       & 886          \\
                                  & SafetyCarCircle-v0            & 300          & 2            & 8           & 100      & 1450         \\
                                  & SafetyDroneCircle-v0          & 300          & 4            & 18          & 100      & 1923         \\
                                  & SafetyAntCircle-v0            & 500          & 8            & 34          & 200      & 5728         \\ \hline
\multirow{9}{*}{MetaDrive}        & SafeMetaDrive-easydense-v0    & 1000         & 2            & 261         & 85      & 1000         \\
                                  & SafeMetaDrive-easysparse-v0   & 1000         & 2            & 261         & 85      & 1000         \\
                                  & SafeMetaDrive-easymean-v0     & 1000         & 2            & 261         & 85      & 1000         \\
                                  & SafeMetaDrive-mediumdense-v0  & 1000         & 2            & 261         & 50       & 1000         \\
                                  & SafeMetaDrive-mediummean-v0   & 1000         & 2            & 261         & 50       & 1000         \\
                                  & SafeMetaDrive-mediumsparse-v0 & 1000         & 2            & 261         & 50       & 1000         \\
                                  & SafeMetaDrive-harddense-v0    & 1000         & 2            & 261         & 85      & 1000         \\
                                  & SafeMetaDrive-hardsparse-v0   & 1000         & 2            & 261         & 85      & 1000         \\
                                  & SafeMetaDrive-hardmean-v0     & 1000         & 2            & 261         & 85      & 1000         \\ \hline
\end{tabular}
}
\end{table}

\subsection{Dataset cost-reward-return plot visualization.}

We visualize the cost-reward-return plot as described in section \ref{sec:preliminary}. Each dot is associated with trajectories with corresponding cost and reward returns. 
Specifically, for each trajectory, we compute its total reward and total cost. Plotting these points on a two-dimensional plane where the x-axis represents the total cost and the y-axis represents the total reward, we obtain a scatter plot that characterizes the trade-offs between reward maximization and constraint satisfaction.

High diversity in a dataset is reflected by a wide spread of points on the cost-reward plot. This implies that the dataset contains trajectories that exhibit various trade-offs between cost and reward, thus providing rich training data for offline safe RL algorithms to learn from. On the contrary, a dataset with low diversity will have points that cluster closely together, indicating limited variety in terms of cost-reward trade-offs.
Such a plot serves as an effective and intuitive tool for understanding the properties of offline safe RL datasets. It offers valuable insights into the dataset's composition, revealing the degree of challenge and diversity embedded within. This, in turn, aids in selecting appropriate datasets for benchmarking and comparison of various offline safe RL algorithms.

\begin{figure}[h]
\vspace{-1mm}
    \centering    \includegraphics[width=0.99\linewidth]{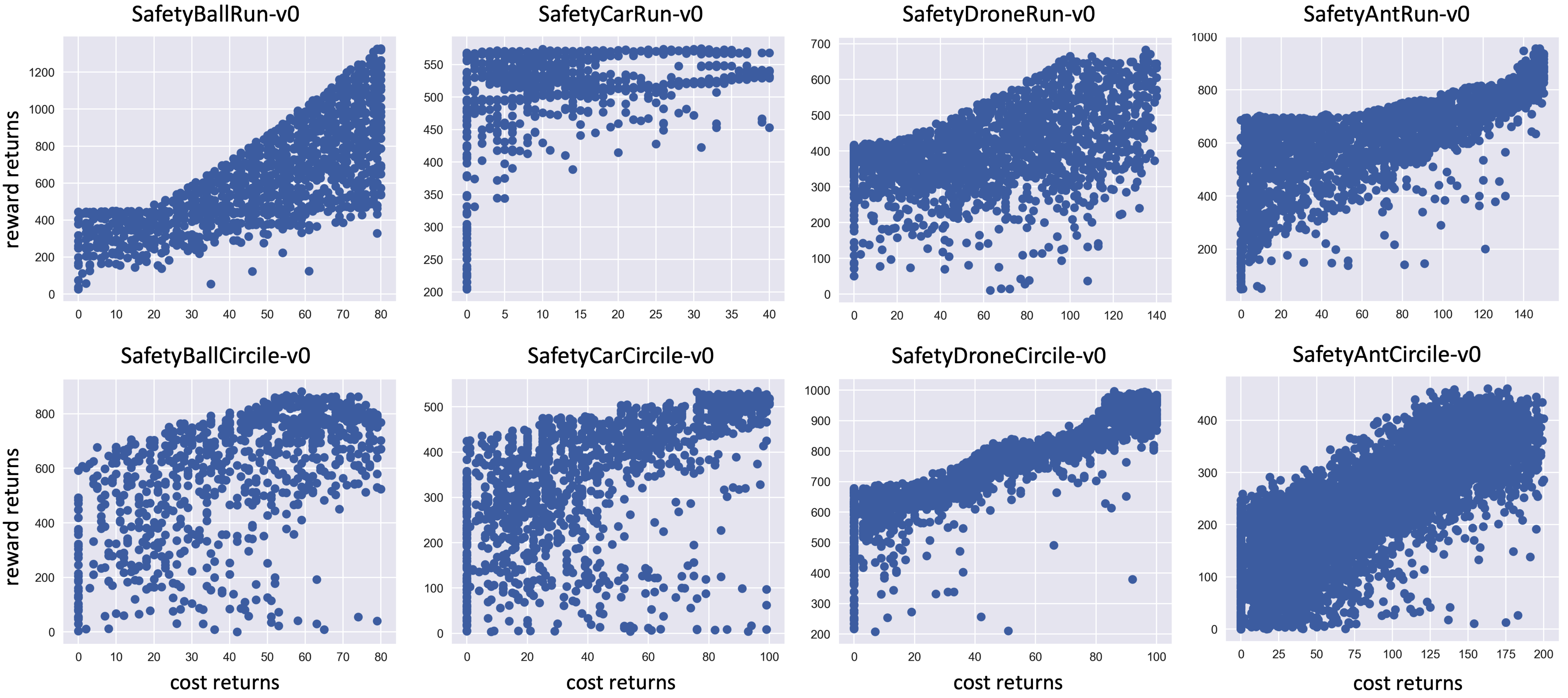}
    \caption{\small Visualization of BulletSafetyGym dataset trajectories on the cost-reward return space.}
    \label{fig:cr-plot}
    \vspace{-1mm}
\end{figure}

\begin{figure}[h]
\vspace{-1mm}
    \centering    \includegraphics[width=0.99\linewidth]{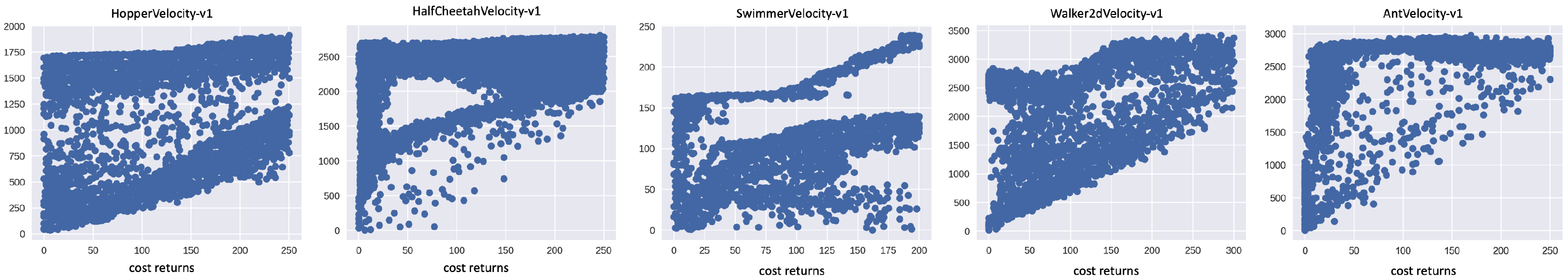}
    \caption{\small Visualization of Velocity dataset trajectories on the cost-reward return space. }
    \label{fig:cr-plot}
    \vspace{-1mm}
\end{figure}

\begin{figure}[h]
\vspace{-1mm}
    \centering    \includegraphics[width=0.95\linewidth]{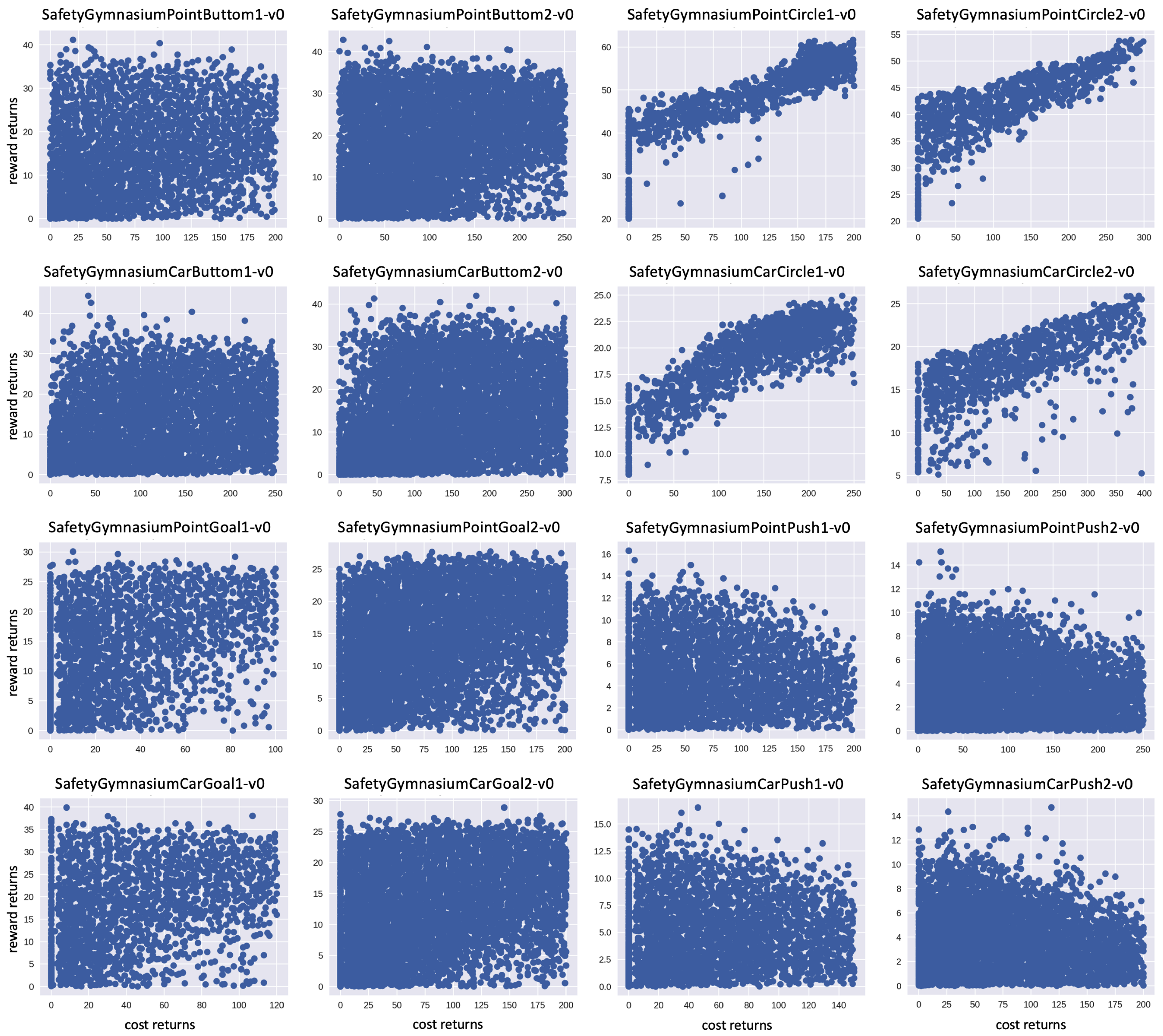}
    \vspace{-2mm}
    \caption{\small Visualization of SafetyGymnasium dataset trajectories on the cost-reward return space.}
    \label{fig:cr-plot}
    \vspace{-2mm}
\end{figure}

\begin{figure}[h]
\vspace{-2mm}
    \centering    \includegraphics[width=0.75\linewidth]{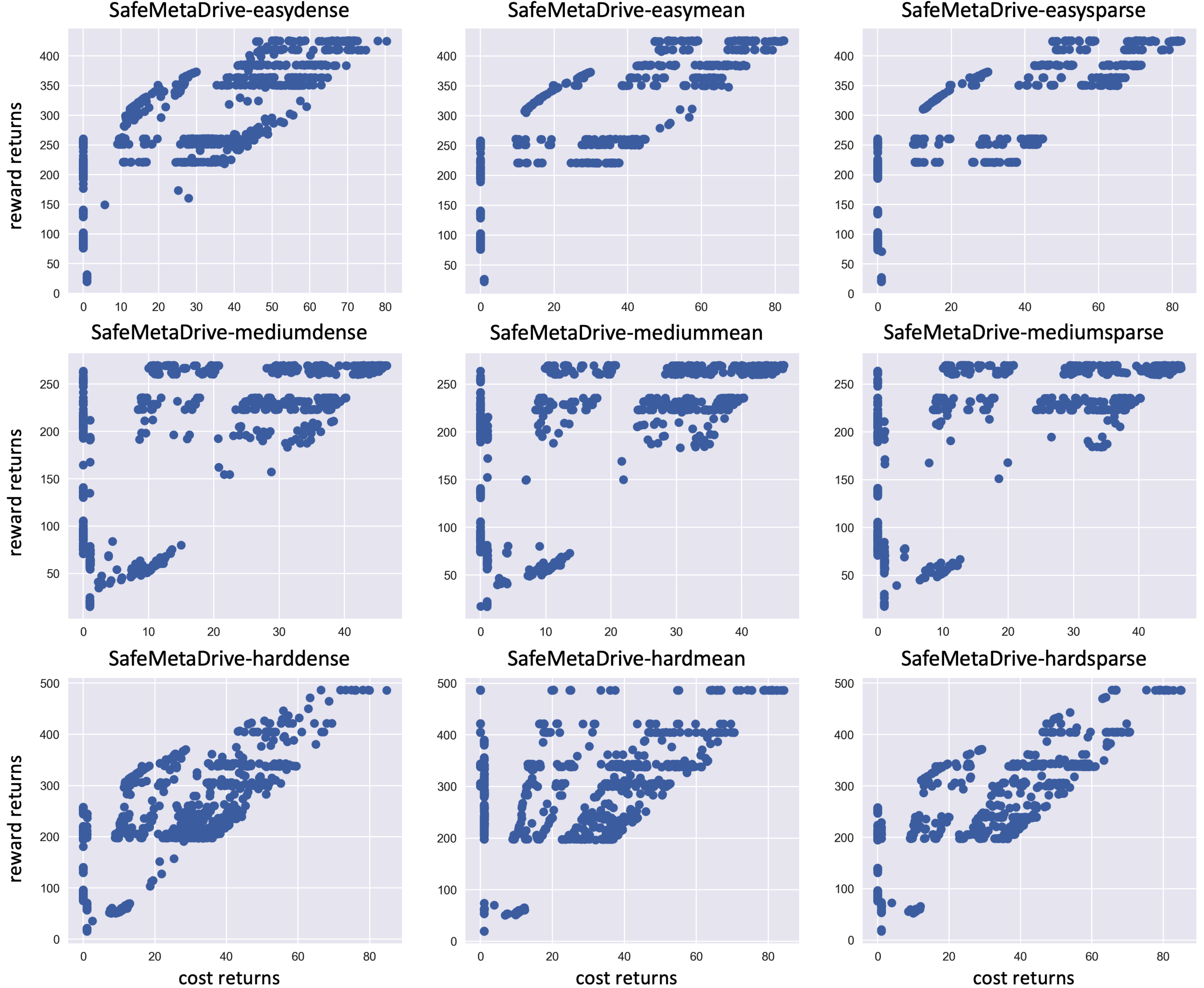}
    \vspace{-2mm}
    \caption{\small Visualization of MetaDrive dataset trajectories on the cost-reward return space.}
    \label{fig:cr-plot}
    \vspace{-2mm}
\end{figure}

Analyzing the figures provided, we can generally discern an increasing trend for the reward frontiers in relation to the cost returns. In other words, as cost return increases, so too might the reward return, underscoring the inherent trade-off between reward and cost. This phenomenon aligns with findings discussed in previous works \cite{liu2022robustness, liu2023constrained}. It's particularly pronounced in BulletSafetyGym tasks and the velocity tasks, as these tasks are largely deterministic - their initial states and transition dynamics are not heavily influenced by randomness. We can thus infer that loosening the safety cost threshold may open up opportunities for task utility reward improvement.

In contrast, the same clear increasing trend is not observable in many highly stochastic SafetyGymnasium tasks, such as Goal, Button, and Push. These tasks introduce an element of randomness in the environment, where the initial state is drawn from a random distribution, significantly impacting the final reward and cost. For instance, in the Goal task, random initialization might result in a direct path between the agent's start position and the goal, enabling the completion of the task with zero constraint violations. Consequently, the datasets contain high-reward, low-cost trajectories due to these "lucky" initializations.

For the autonomous driving tasks in MetaDrive, the cost results from three safety-critical scenarios: (i) collision, (ii) out of road, and (iii) over-speed. In this case, the environment's stochasity mainly comes from the random initialization of surrounding traffic flows and the map configuration. To foster the diversity of sampled trajectories within offline datasets, we utilize varying parameters to moderate the aggressiveness of the Intelligent Driver Model (IDM) policies~\cite{kesting2007general} of the ego vehicles.  
We can also observe an increasing trend for the reward frontiers with respect to the episodic cost returns in most of these environments. 

It's worth noting, however, that even though the cost-reward return plot of the dataset might not accurately reflect the reward-cost trade-off, the training curves of the expert policies do display a significant trend. This is because each policy is evaluated on multiple episodes and uses expectations as the evaluation metrics. In other words, under varying cost conditions, the cost value function and the reward value function of the policy can still reflect the trade-offs when considering expectations. This concept is discussed in more detail in \cite{liu2022robustness}.

\end{document}